\pdfoutput=1
\documentclass[11pt]{article}

\usepackage[final]{acl}

\newcommand{\llama}{Llama 2 }

\usepackage{times}
\usepackage{latexsym}
\usepackage{float}
\usepackage{subcaption}
\usepackage{booktabs}

\usepackage[T1]{fontenc}
\usepackage[utf8]{inputenc}

\usepackage{microtype}

\usepackage{inconsolata}

\usepackage{todonotes}
\usepackage{graphicx}
\usepackage{amsmath,amsfonts,amssymb}
\usepackage[most]{tcolorbox}
\usepackage{subcaption}
\usepackage{cleveref}

\title{Are Female Carpenters like Blue Bananas? A Corpus Investigation of Occupation Gender Typicality}

\author{Da Ju \and
  Karen Ulrich \and
  Adina Williams \\
  FAIR Laboratories\\ Meta Platforms, Inc.\\
  \{\texttt{daju, karenu, adinawilliams}\}\texttt{@meta.com} \\}

\begin{document}
\maketitle
\begin{abstract}
People tend to use language to mention surprising properties of events: for example, when a banana is blue, we are more likely to mention color than when it is yellow. This fact is taken to suggest that yellowness is somehow a typical feature of bananas, and blueness is exceptional. Similar to how a yellow color is typical of bananas, there may also be genders that are typical of occupations. In this work, we explore this question using information theoretic techniques coupled with corpus statistic analysis. In two distinct large corpora, we do not find strong evidence that occupations and gender display the same patterns of mentioning as do bananas and color. Instead, we find that gender mentioning is correlated with femaleness of occupation in particular, suggesting perhaps that woman-dominated occupations are seen as somehow ``more gendered'' than male-dominated ones, and thereby they encourage more gender mentioning overall. 
\end{abstract}

\section{Introduction}\label{sec:intro}

\begin{figure}[ht]
  \centering
  \includegraphics[width=\linewidth]{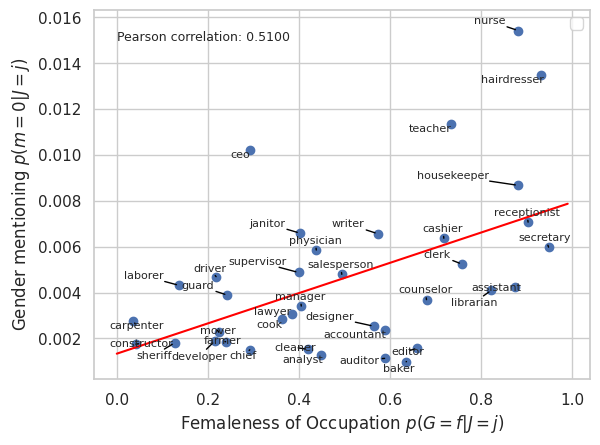}
  \caption{We found the strongest correlation between the femaleness of an occupation (according to US labor statistics) and gender mentioning in \textbf{Pushshift.io Reddit}, a surprising finding to some extent, because it contradicts the idea that gender mentioning occurs when special events are being pointed out. Instead, this finding points more to a gender-specific phenomenon.}
  \label{fig:page1plot}
\end{figure}
 
% Gender can take many forms in language. One of the most interesting cases is occupation nouns:  % For example, the occupation noun ``carpenter'' is commonly used to refer to men. Of course, it is not a necessary property of carpenters that they be men, but, currently in the US, for example, the majority of carpenters are. 
While people of any gender can do any job, most occupations in the current world are not fully gender-balanced, meaning a statistic association exists between particular genders and particular occupations. We can see evidence of such associations in how humans talk about occupations. When talking about a carpenter who is a woman---a male-dominated profession in the US---people may refer to her as ``female carpenter''. This usage suggests that a woman being a carpenter is rare (and potentially surprising), and thus some find it worth remarking upon explicitly. Men on the other hand, are more likely to be called ``carpenter'' instead of ``male carpenter'', presumably because the latter feels redundant or overinformative somehow. %This usage, if predominant in a corpus, might indicate that a woman being a carpenter is comparatively rare (and potentially surprising), and thus some people find it worth remarking upon. On the other hand, men being carpenters is quite common and, by a similar token, 
%The phrase ``male carpenter'' (i.e. ``male mentioning'') probably only occurs in corpora in situations where someone wants to refer to the only man present amongst a set of mostly non-men carpenters.

Typicality effects such as these shape the way we use language in general. For example, \citet{Degen2020redundancy} adduced evidence for typicality effects in the color domain, finding that speakers never referred to an image of a lone yellow banana as ``the yellow banana'', but almost always referred to a lone blue banana as ``the blue banana''. Because we think of the (stereo-)typical banana as yellow, the blueness of a blue banana is more worth remarking upon than the yellowness of a yellow one. In this work, we explore a parallel phenomenon in the occupation gender space: Are there gender typicality effects for occupation nouns? If so, we expect more mentions of ``male'' when an occupation is woman-dominated, and more mentions of ``female'' when an occupation is man-dominated. % This could tell us how much a particular gender is ``typical'' of an occupation (based on the empirical corpus distribution of occupation words and gender mentions).

Alternatively, gender mentioning may occur more often when the gender associated with an occupation is salient. Past work in social psychology and gender studies has suggested that gender is generally more salient for women than for men. In particular, men are seen as more typical of the concept of ``person'' than women are, and that women are more associated with gender than men are \citep{gilman1911man, hamilton1991masculine, bem1993lenses,merritt1995attribution, bailey2019man, bailey2020implicit, bailey2022based, bailey2024intersectional}. Put simply, according to these theories, the concept of `woman' differs from the concept of `man' in that it has an additional semantic attribute that encodes cultural and/or biological facts or connotations (such as information about clothing, behaviors, etc.) that are associated with women. 

In light of this past work, it is also plausible that associating women with ``having gender'' affects how we conceptualize woman-dominated occupations, such as ``nurse'' or ``hairdresser''. If woman-dominated occupations are indeed conceptualized as ``more gendered'', then encountering a woman-dominated occupation might make gender itself more salient, and lead to an increase gender mentioning in corpora. Were this to be the case, we would see higher overall mentioning of gender (with both ``male'' and ``female'' adjectives) for woman-dominated occupations.

% Alternatively, peoples' impression of the genderedness of particular occupations may not be directly related to the actual empirical gender breakdown of the occupations. Perhaps people mention gender more for occupations that are highly gender-coded. There is some evidence for this from social psychology work which explores the androcentrism hypothesis, which states that masculine is the default, and seen as more prototypical of humans \citep{}. Relatedly there may There are some occupations, such as ``physician'' for which the masculine gender is seen as more likely, but in point of fact, the gender balance of the existing population is fairly balanced. For these occupations, we might wonder whether the gender mentioning  
In this study, we test these two hypotheses about occupation gender typicality using information theoretic approaches to analyze large text corpora.  
First, we set aside the null hypothesis that gender mentioning (modifying an noun with either ``male'' or ``female'' adjective) is completely unrelated to occupation gender (i.e., the empirical gender breakdown in the real population), see Section~\ref{sec:null}.

Next, we test the hypothesis that gender is more salient for women-dominated occupations. This hypothesis is borne out: We find that gender (both male and female) is mentioned more when the occupation is woman-dominated in our largest tested corpus, see Section \ref{sec:femalesalient}. We report medium-to-large correlations between femaleness of occupation (the degree to which it is woman-dominated, according to U.S. labor statistics) and overall gender mentioning ($r=0.49$), and for femaleness of mentioning ($r=0.50$), maleness of mentioning ($r=0.42$), respectively. 

Finally, we test whether gender mentioning correlates with surprisingness, or non-typicality, of gender given an occupation. To test this, we define an information theoretic quantity based on the conditional entropy of female/male gendering given a particular job, whereby low occupation genderedness would correspond to a 50-50 gender breakdown. This can be seen as testing the ``blue banana'' hypothesis for occupation gender typicality, see Section~\ref{sec:bluebanana}. We do not find strong evidence for this hypothesis, with very weak correlations emerging for two different source datasets, Pushshift.io Reddit ($r=0.13$) and Wikipedia ($r=-0.12)$. 

In sum, our results support the idea that occupation gender typicality is related to gender mentioning, but not in the way predicted by the ``blue banana'' hypothesis. Instead, gender mentioning appears to be more related to gender salience for woman-dominated occupations, at least for the the corpora we investigated. 

To further analyze our results, we also explore three secondary questions. First, we ask: does corpus choice matter (see Section~\ref{sec:corpusresults}), and find that our effects are replicated in both Wikipedia and Pushshift.io Reddit, but that they are stronger for the Pushshift.io Reddit dataset. Second, we ask: do the empirical estimates of gender breakdown actually reflect the way people perceive the genderedness of occupations (see Section~\ref{sec:coded}).
Utilizing a standard word-embedding based method to determine perceived gender, we find that, while similar, perception and empirical observations do meaningfully differ. Third, we ask whether there are trends in the meaning of sentences which mention gender and occupations, such as negative sentiment (see Section~\ref{sec:sentiment}).% In addition, we also explore the effects of regenerating data using a language model (LM). To do this, we prompt a large open language model with input strings from Wikipedia and collect the output as a new text source. We measure the extent to which occupation gender typicality has changed throughout this process and find the LM generated Wikipedia to approximate the findings from the original Wikipedia. We also observed that the LM-regenerated Wikipedia dataset has the weakest correlation between femaleness of occupations and gender mentioning, suggesting that the LM actually lessened the association between gender mentioning and the femaleness of woman-dominated occupations to a small degree. 

%As an additional analysis, we explore the extent to which the empirical estimates of gender breakdown actually reflect the way people perceive the genderedness of occupations. Utilizing a standard word-embedding based method to determine perceived gender, we find that, while similar, perception and empirical observations do meaningfully differ. We then show that the perceived occupation gender can be used as further evidence that mentioning is correlated with occupation femaleness.

\section{Terminology and Notation}

\subsection{Terminology}
Gender is a complex social phenomenon that manifests in how we use language and operate in our societies. Here, we draw upon a distinction described in \citet{ackerman2019syntactic} between social gender, which is how people see themselves in the world, and conceptual gender, which refers to how lexical items are associated with masculine or feminine properties, in the absence of explicit, formal grammatical features such as grammatical gender affixes, gender concord etc. (c.f. \citealt{hoyle-etal-2019-unsupervised, williams-etal-2019-quantifying, williams-etal-2021-relationships}). When we say ``gender'' in this work, we will be referring to conceptual gender. 

Since occupation words can't encode the social genders of individuals in English (because English doesn't use grammatical gender marking systematically on all nouns), we decided to use the terms ``female'' and ``male'' here.\footnote{We use these terms instead of ``feminine'' and ``masculine'', because we want to avoid confusing occupation gender with the morphological categories of grammatical gender.} We will use the terms ``woman'' and ``man'' when referring to the social genders. For the most part, ``man'' and ``woman'' will be used for illustrative examples, or when we discuss the survey data from the U.S. Bureau of Labor Statistics, which is based on self-report. 

In what follows, we have carved out a particular notion of gender which enables us to test existing hypotheses about the gender typicality of occupations. To do this, we have had to code gender as a binary variable. Clearly, this is not empirically sufficient to attain full coverage of all humans, as there are many genders in the English-speaking world, but we are restricted to this coding based on the availability of existing resources, see Section~\ref{subsec:USStats}. Given our precise research questions, the restrictions we have inherited from existing resources do not overly limit or negatively impact our conclusions, as currently, we are aware of no occupations that appear to be ``typical'' of non-binary individuals or those with other genders. The lack of occupations typical of non-binary people is probably due to the small number of people who currently identify as non-binary, which are estimated to range from approximately 0.68\% \citep{wilson2021nonbinary} to 1\%\footnote{\href{https://www.pewresearch.org/short-reads/2022/06/07/about-5-of-young-adults-in-the-u-s-say-their-gender-is-different-from-their-sex-assigned-at-birth/ft_2022-06-07_transandnbadults_01/}{Pew Research Center, 2022}} of the US population. This may change in the future, and we look forward to future research exploring the nature of this change.  

\subsection{Notation}\label{subsec:notation}
We describe our notational conventions with respect to random variables, entropy, and mutual information in the Appendix~\ref{appdendix:notation}.

We specifically define three r.v.s in the context of this study; 
the occupation noun associated with a person $J$, the gender associated with the occupation noun $G$, and if that gender was mentioned as a descriptor $M$.
$M$ and $G$ are assumed to have a binary alphabet in the context of this investigation:  $\mathcal{M}=\{$`mention', `no mention' $\}$, and  $\mathcal{G}=\{$`female identifier', `male identifier' $\}$, if appropriate we may also use $\{0,1\}$ as shorthand notation for both. The alphabet for $J$ is a list of 37 occupation nouns that we have specifically listed in Appendix~\ref{sec:listofoccupation}.

For example ``female physician" is associated with $(j=$`physician', $m=$`mention', $g=$`female identifier'$)$. Note that when $m=$`no mention', we can not determine $g$. Hence, we do not have access to the full joint distribution $P_{JMG}$. In Section~\ref{sec:method}, we will show how we can make statements about gender disparities, despite not being able to observe outcomes for $g$ in many cases.

% Further, note that obviously, there are more than two genders in the English-speaking world, however, references to non-women, non-men are not sufficiently prevalent in our corpora (or in the Labor Statistics) to yield a reasonable estimate in the corpora available (see next section). We would be very interested to incorporate the full range of existing genders in the future.
\section{Data Sources}
We use four different data sources for this work, which we discuss in turn below. 

To address our main research questions, we first find an empirical estimate of occupation gender (from the US Bureau of Labor Statistics). Then, we select two source corpora on which to run our analysis: We use Wikipedia and Pushshift.io Reddit, because they are extensive, openly available resources, which are commonly used in the NLP field, and frequently as training data for Language Models Language Models (LMs). Both should ensure a reasonable coverage of occupations with gender mentions for our analysis. 

Finally, we prompt the opensource \llama language model \citep{2023-llama2} to regenerate text in the style of Wikipedia. We call this new dataset \llama Wikipedia and use it to investigate whether occupation gender typicality measurements change as a result of a corpus being rewritten by an existing language model. 

\subsection{US Labor Statistics}\label{subsec:USStats}
Following \citet{caliskan-etal-2017-semantics,rudinger-etal-2018-gender, zhao-etal-2018-gender, bartl-etal-2020-unmasking, gonzalez-etal-2020-type} i.a., our empirical estimates of occupation gender breakdowns have been obtained from the U.S. Bureau of Labor Statistics\footnote{\url{https://www.bls.gov/cps/cpsaat11.htm}}. These estimates have been collected monthly by the Bureau of Census on behalf of the Bureau of Labor Statistics based on household survey. This survey, the Current Population Survey provides statistics about worker demographics, among other things, where individuals self-report their (binary) genders and occupations. We utilize labor statistics for 2023 (accessed in late fall).% One limitation of this resource is that gender is coded as binary, as opposed to covering the full rage of existing genders. 

We note that some occupations, such as `attendant' are aggregated to include all roles with `attendant' in their titles, including flight attendant, parking attendant, transportation service attendant, dining room and cafeteria attendants and bartender helpers. In some cases, occupations are missing from the statistics, so we substitute them with synonyms such as ``Chief Executives" for ``CEO'' or ``Police Officers" for ``sheriff''. The comprehensive aggregated statistics are available in Appendix \ref{sec:comparison}.

\subsection{Pushshift.io Reddit}
Building upon the work of \citet{humeau2020polyencoders}, we employ a pre-established dataset sourced from the online social forum, Reddit. This dataset, acquired and made available by an independent third party, can be accessed publicly on \texttt{pushshift.io} \citep{Baumgartner_Zannettou_Keegan_Squire_Blackburn_2020}. It encompasses data collected from PushShift\footnote{https://files.pushshift.io/reddit/} up to July 2019.
We follow a similar data cleaning approach to \citet{roller2020recipes} to enhance the clarity of the signal. 
A comment, along with all its subsequent child comments, is eliminated if it fulfills any of the subsequent criteria: 1) The author's identifier includes the term `bot'; 2) The origin of the comment is a known non-English Subreddit; 3) The comment is flagged as removed, deleted, or is devoid of content; 4) The comment comprises less than 70\% alphabetic characters; 5) The comment contains a URL. 
% Contrary to \citet{roller2020recipes}, who split threads into multiple examples for training, leading to comment repetition, our study adopts a different approach. We aim to analyze gender and occupation distribution in the Reddit dataset. 
To avoid comment repetition, we flatten each thread tree by concatenating comments in a pre-order traversal sequence. The processed data contains 301M threads. 

\subsection{Wikipedia}\label{subsec:Wiki}
Wikipedia is a fairly large datasource containing encyclopedic information in many languages, which is moderated by volunteers. In conducting our research, we utilize an English Wikipedia dump dated April 2023. The dataset is prepared in accordance with the procedures outlined by \citet{Wikiextractor2015}. This extracts articles and associated metadata from the raw XML files. Despite Wikipedia's general suitability to our research questions, it is worth noting that Wikipedia data tends to be stereotypically skewed towards men, as highlighted by \citet{wagner2015its, schmahl-etal-2020-wikipedia, falenska-cetinoglu-2021-assessing}.

\subsection{Linguistic Analysis}
Upon obtaining the pre-processed datasets, we adhere to the pipeline delineated by \citet{williams-etal-2021-relationships}. For this endeavor, we utilize the Stanza tool \cite{qi2020stanza}. The pipeline comprises the following steps: 
1) Tokenization \& Sentence Segmentation; 2) Part-of-Speech (POS) Tagging: Each token is assigned grammatical information (e.g., noun, verb, adjective); 3) Lemmatization: Each token is reduced to its base form, facilitating the normalization and simplification of the text; 4) Dependency Parsing: This final phase involves analyzing the grammatical structure of  each sentence, thereby establishing relationships between words, as described in \citet{chen-manning-2014-fast}. The size of the datasets, following this procedure, is detailed in Table~\ref{tab:datasetsize}.

\begin{table}[!ht]
\centering
\begin{tabular}{lc}
\toprule
\textbf{Dataset} & \textbf{Size}\\
\midrule
Pushshift.io Reddit & 118.81 billion \\
Wikipedia & 3.01 billion \\
\llama Wikipedia & 2.98 billion\\ 
\bottomrule
\end{tabular}
\caption{Total size of datasets in (word) token count.}
\label{tab:datasetsize}
\end{table}
%
%\vspace{-12mm}
\subsection{Extraction of gender - occupation pairs}
\label{subsection:genderextraction}

Initially, our task involves extracting all occupation-related words from our text resource and isolating the subset that includes gender mentions. To accomplish this, we commence with a basic list of occupation nouns, sourced from the U.S. Bureau of Labor Statistics (Appendix \ref{sec:listofoccupation}). Additionally, we utilize a concise list of gender adjectives, namely ``male'', ``female'', ``masculine'', ``feminine'', ``man'', ``woman'', "non-binary'' and ``nonbinary''.\footnote{We find very few occupation nouns modified with ``non-binary'' in Wikipedia (40) and Pushshift.io Reddit (101), therefore exclude them from the analysis for lack of signal.}

Our approach involves identifying instances where an occupation noun appears in the text, followed by determining when a gender adjective modifies this noun. We then extract all \texttt{amod} dependencies, where an occupation noun serves as the head and a gender adjective as the dependent. In this way, we isolate all gender mentions and further categorize them by gender.
The statistics pertaining to occupation mentions and gender-specific mentions derived from these pairs can be found in Tables~\ref{appendix:reddittable}--\ref{appendix:llamatable}.

\subsection{\llama Wikipedia}

Recent work has found that LMs can exacerbate gender imbalances present in their training data \citep{kotek-etal-2023-gender}. To investigate the extent to which LMs generate text that statistically matches what we find for occupation gender typicality, we investigate whether results would change if we examined a different version of Wikipedia, specifically one regenerated by an LM. If the results on LM-restructured Wikipedia are substantially different, it would suggest that the LM either introduces new gender bias, or removes existing gender bias.  
We utilize \llama 70B to generate Wikipedia content and subsequently analyze the distributional differences in occupation gender mentions. We confine our experiments in LM-generated datasets to Wikipedia, as Pushshift.io Reddit is too large to be tractable. 

To generate new text, we provide the first 256 words from the original Wikipedia article, along with the article's title, from the original content describe above in Section~\ref{subsec:Wiki}. We instruct the model to produce an article of a minimum length of 700 words, aligning with approximately the average English Wikipedia article length of 658 words. The prompt used is in Appendix~\ref{sec:promptllamawikipedia} to enable replicability of our results. We apply the same data processing and extraction procedure as for the original Wikipedia dataset. From Table~\ref{tab:datasetsize}, we can observe that the two are comparable in size.

We evaluate the BLEU score, a text similarity metric, between the original Wikipedia articles and those generated by Llama 2. Our findings reveal low BLEU scores (as detailed in Appendix \ref{appendix:bleuscore}), indicating substantial textual differences between the original Wikipedia content and the \llama regenerated versions. This suggests that the \llama Wikipedia dataset warrants statistical analysis.

\section{Methods}\label{sec:method}

In this section, we will first describe how observed occurrence counts translate to probability mass functions (pmfs) relating to occupation, gender mentioning and gender.
Henceforth, we describe how to compute quantitative markers for gender mentioning and gendered occupations, which we finally use to test our hypotheses from in Section~\ref{sec:intro}.

\subsection{Estimation of probability mass functions}

The US Labor Statistics is our source to compute the the relationship of gender and occupation independent of gender mentioning in language.
Specifically, we compute $P_{G,J}$ by dividing the number of men / women employed in a specific occupation by the total number of employees in the survey.
This distribution will be used as the foundation to determine how statistically biased an occupation is with respect to gender. 

For each text source, we first determine the likelihood of gender mentioning independent of the specific gender $P_{M}$. It is the number of gendered occupation mentions divided by the total number of occupation mentions. 
We further compute, the conditional joint distribution $P_{G,J|M=0}$. Similarly to the US Labor Statistics, we divide the number of male / female indicated occurrences of a specific occupation by the total number of gendered observations. Note that it would be more informative if we could compute $P_{G,J,M}$ directly, however since we can not determine gender if it is not mentioned, we need to approximate the joint distribution by combining information from the text source and US Labor Statistics.

%To illustrate the process. Consider the following example, we count the occurrence of job descriptions in Wikipedia and record the gender of the person if it was modified with a gender adjective to compute $P_{J,G|M=0}$ and $P(M)= \sum_{M\in \mathcal{M}} P(J,M)$. For example, say the word "physician" 100 times, 50 times a gender adjective was used, 45 times 'male' and 5 'female'. As we can only measure the gender distribution from a text directly when it is mentioned, we must find a proxy for the gender distribution when it was not mentioned. We cannot assume that the two distributions will match or that the gender distribution will be 50/50 (e.g., hairdressers are nearly all women in reality but adjective mentions of hairdresser gender are skewed to a lesser degree towards female, see table below). 

\subsubsection{Computing the joint}
To compute $P_{G,J,M}$, we need the previously computed pmfs;  $P_{G,J}$, $P_{M}$ and $P_{G,J|M=0}$, where the latter two are computed from the text corpus, and the the first one is computed via the US Labor Statistics.
By applying Bayes' theorem and the chain rule of probability, we can compute the joint likelihood as follows; 
\begin{align} 
P_{G,J,M=0} &=  P_{G,J|M=0}\cdot P_{M=0} \\
P_{G,J,M=1} &= \frac{P_{G,J}-P_{G,J,M=0}}{P_{M=1}} \label{eq:bayes} \\ 
P_{G,J,M}   &= P_{G,J|M}\cdot P_{M}                             \label{eq:chainrule}
\end{align}

Note, that we assume that there exists a joint likelihood for which the US Labor Statistics is the correct marginal $P_{G,J} = \sum_m P_{G,J,M=m}$, and our text corpus gives the correct conditional $P_{G,J|M=0}$. This assumption can be a problem, when there is a statistical bias in what occupations are more prominent in a text sources as opposed to the labor statistics. This can be problematic for multiple reasons: (1) some occupations such as ``author" will be over-represented in a text source such as Wikipedia, and (2) the gender and job distribution of a text source and labor statics needs to match, which can easily not be true because either comes from a different cultural or temporal context.
However, in our analysis especially for Pushshift.io Reddit, we believe there to be a good match, as Reddit was created in 2005 and sampled in 2019, and the labor statistics have in all likelihood not changed dramatically since then.

\subsection{Gender bias in occupations}

We generally cannot access how people perceive the gender typicality of an occupation directly from text, but we can hope their perceived typicality matches the real world gender breakdown\footnote{For words such as ``manager" this might not always be correct.}.
Consequently, in this study, we propose to measure \textbf{occupation genderedness}, which we define as 
\begin{align}
    1 -H(G|J=j). 
\end{align}
Note that, according to our definitions in Appendix \ref{appdendix:notation}, the entropy term will be in the range between 0 and 1. 
Consequently, occupation genderedness is low for unbiased occupations and high for gender biased occupations (i.e. man-dominated or women-dominated occupations). 
We further compute the \textbf{femaleness of occupation};
\begin{align}
    P_{G=0|J=j}.
\end{align}
Both are computed from the US Labor Statistics joint distribution $P_{G,J}$.

We introduce both of these measures to test different hypotheses.
The former measure is a symmetric estimator. 
If there is no gender specific surprise, we should find occupation genderedness to correlate well with gender mentioning, In other words we expect ``male nurse'' and ``female CEO'' to both be likely expressions. 
The second measure allows for detecting gender specific mentioning.

%For the purposes of this note, we will call occupation words like "physician" biased, in the sense that a gender adjective is present for these jobs (see table below). 

\subsection{Gender mentioning in text}

We measure two ways gender is mentioned in text: by computing (1) $P_{M=0|J=j}$ directly for the likelihood of any gender being mentioned, $P_{M=0|J=j,G=0}$ for femaleness being mentioned,  $P_{M=0|J=j,G=1}$ for maleness being mentioned, and (2) the mutual information between gender and its mentioning $MI(G;M|J=j)$.

We can estimate the correlation strength of gender bias markers and gender mentioning markers by computing the Pearson correlation coefficient of a linear regression. 
Further, as discussed in \citep{caliskan-etal-2017-semantics}, we can make group comparisons of mutual information between female and male occupations based on distance between text embeddings. The technical details can be found in Appendix \ref{sec:femcodewomendom}.
% \todo{DExter: Can you put what you did here???!}

%If our hypothesis that speaker surprise is the main predictor for mentioning gender identifiers holds true, we expect to be able to show two correlations:
%(1) We hypothesis that gender bias in occupation predicts the likelihood of mentioning gender $p(M=`mention'|J=j)$. 
%(2) And we expect mutual information $I(G;M|J=j)$ to be  correlated as well.
% And finally, we measure the mutual information $MI(G;M|J=j)$. As $M$ and $G$ are binary random variables, $P(M|J=j)$ and $P(G|J=j)$ are considered unbiased if they have likelihood $0.5$.
%This would exclude not just an uncorrelated usage of gender identifiers but also the hypothesis that in unbiased occupations, genders is mentioned because it would carry much information (informative only).

\begin{figure*}[ht]
  \centering
  \includegraphics[width=.75\linewidth]{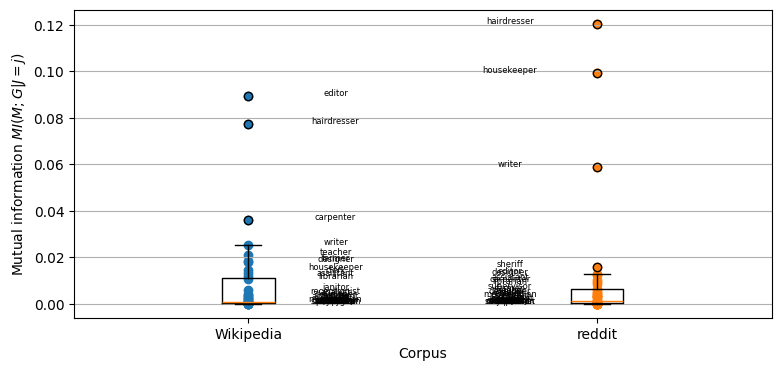}
  \caption{We calculate the conditional mutual information between gender and its mentioning  by means of a gender indicator $MI(G;M|J=j)$. Very low mutual information indicates that  variables are not correlated.
  We see a wide spread in MI across occupations $j$. However, we see similar occupations in the top spots for Wikipedia and Pushshift.io Reddit. }
  \label{fig:mutual_information}
\end{figure*}
\begin{figure*}[ht]
  \centering
  \begin{subfigure}[b]{0.3\linewidth}
    \centering
    \includegraphics[width=\linewidth]{images/e.png}
    \caption{}
    \label{fig:plot1}
  \end{subfigure}
  \hfill
  \begin{subfigure}[b]{0.3\linewidth}
    \centering
    \includegraphics[width=\linewidth]{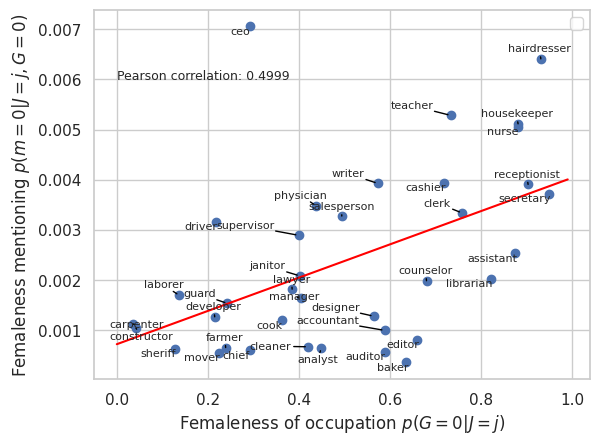}
    \caption{}
    \label{fig:plot2}
  \end{subfigure}
  \hfill
  \begin{subfigure}[b]{0.3\linewidth}
    \centering
    \includegraphics[width=\linewidth]{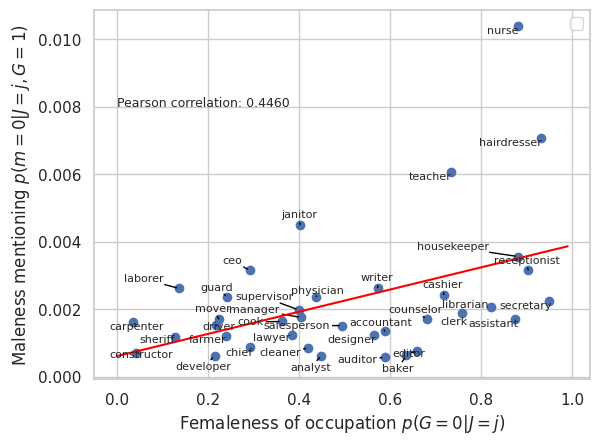}
    \caption{}
    \label{fig:plot3}
  \end{subfigure}
  \caption{We found correlations between the femaleness of an occupation (according to US labor statistics) and (a) gender, (b) femaleness, (c) maleness mentioning in \textbf{Pushshift.io Reddit}.}
  \label{fig:salientgender}
\end{figure*}

\section{Results}

\subsection{Rejecting the Null Hypothesis}\label{sec:null}
First, we test whether we can reject the null hypothesis that gender mentioning is unrelated with occupation genderedness. 
For this, we calculate the conditional mutual information, as described in Section~\ref{subsec:notation}. If the null hypothesis is true, if there is no correlation, and we expect $MI(M;G|J=j) = 0$. We present our results in Figure~\ref{fig:mutual_information}, with individual occupation noun results in Appendix~\ref{sec:miredditwiki}. These findings are compatible with the existence of relationship between gender mentioning and occupation genderness, and the next two sections will attempt to determine its nature.

\subsection{Mentioning gender when it's female}\label{sec:femalesalient}
Next, we explore the hypothesis that gender mentioning is correlated with femaleness of occupation $P_{G=f|J=j}$, under the assumption described in the introduction whereby woman-dominated occupations make gender more salient. We find that this is borne out for Pushshift.io Reddit, see Figure~\ref{fig:salientgender}. We find significant correlations of moderate size between the femaleness of an occupation and mentioning of either gender ($r=0.51$), mentioning of femaleness ($r=0.50$), and mentioning of maleness ($r=0.45$). Since this finding is robust across all three kinds of mentioning, we take this to be evidence in favor of the hypothesis that the female genderedness of woman-dominated occupations makes gender more salient in general, leading to more mentions of gender.

For Wikipedia, we do not find a strong effect of gender mentioning when female, see Figure~\ref{fig:salientgenderwiki}. This effect is compatible with \citet{wagner2015its} and \citet{schmahl-etal-2020-wikipedia}, which find evidence, respectively, for linguistic gender bias in Wikipedia based on adjective sentiment and topic modeling or word embedding approaches. Our findings pertain to gender mention and occupation gender typicality, not to negative sentiment about women or whether stereotypes about family and science exist in the data source---both can be true simultaneously. 
\begin{figure*}[ht]
  \centering
  \begin{subfigure}[b]{0.45\linewidth}
    \centering
    \includegraphics[width=\linewidth]{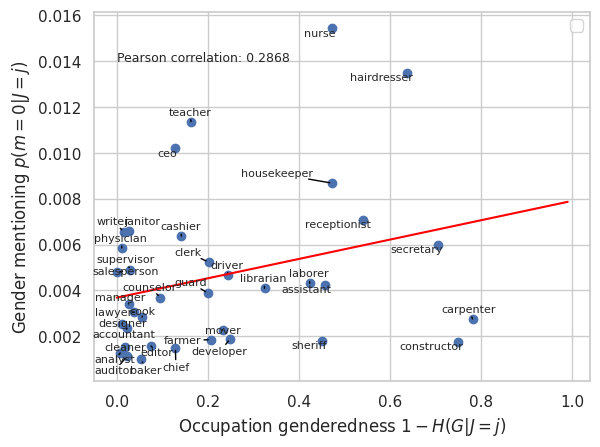}
    \caption{Pushshift.io Reddit}
    \label{fig:plot1}
  \end{subfigure}
  \hfill
  \begin{subfigure}[b]{0.45\linewidth}
    \centering
    \includegraphics[width=\linewidth]{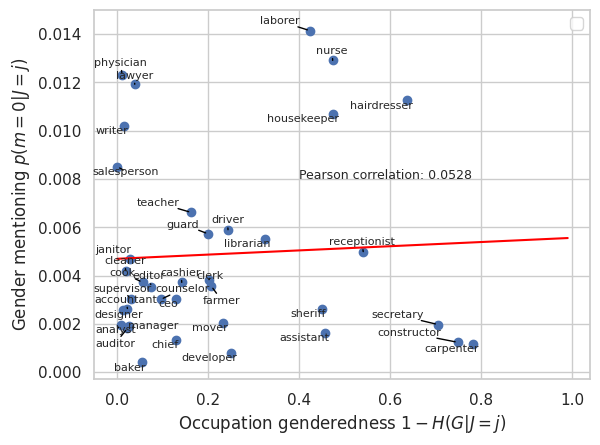}
    \caption{Wikipedia}
    \label{fig:plot2}
  \end{subfigure}
  \caption{Ablation: We tested the correlation of occupation genderness $1 - H(G|J=j)$ and gender mentioning. High occupation genderedness implies either a man- or woman-dominated occupation according to US labor statistics. Observed correlations are weak, eliminating the hypothesis that gender mention is a result of surprise.}
  \label{fig:ablation}
\end{figure*}
\begin{figure*}[ht]
  \centering
  \begin{subfigure}[b]{0.3\linewidth}
    \centering
    \includegraphics[width=\linewidth]{images/e.png}
    \caption{Pushshift.io Reddit}
    \label{fig:plot1}
  \end{subfigure}
  \hfill
  \begin{subfigure}[b]{0.3\linewidth}
    \centering
    \includegraphics[width=\linewidth]{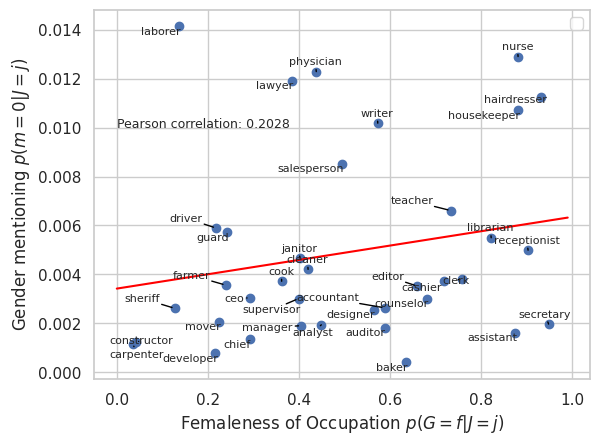}
    \caption{Wikipedia}
    \label{fig:plot2}
  \end{subfigure}
  \hfill
  \begin{subfigure}[b]{0.3\linewidth}
    \centering
    \includegraphics[width=\linewidth]{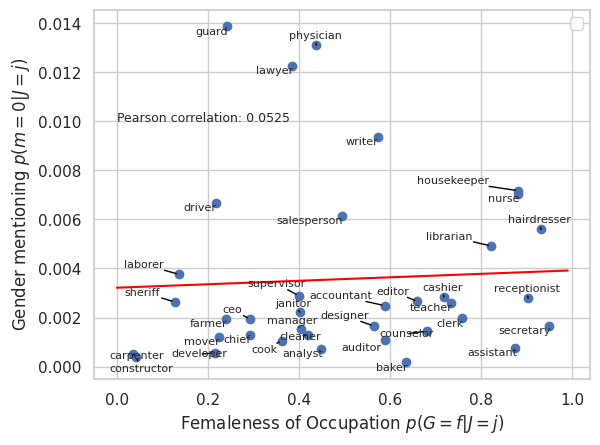}
    \caption{LLama Wikipedia}
    \label{fig:plot3}
  \end{subfigure}
  \caption{Corpus comparison: The femaleness of occupation is most strongly correlated with gender mentioning in Pushshift.io Reddit. In Wikipedia, the effect is smaller, and interestingly, it keeps diminishing for \llama Wikipedia. }
  \label{fig:gender_mentioning}
\end{figure*}

\subsection{Ablation: Female Carpenters are not like Blue Bananas}\label{sec:bluebanana}

We also explored the ``blue banana'' hypothesis: the gender which is not typical of the occupation will be mentioned. We defined occupation genderedness above as $1-H(G=g | J=j)$, where low occupation genderedness meant 50-50 gender breakdown, or gender balance, and high occupation genderedness meant a strongly woman- or a man-dominated occupation. Our results found practically no correlation for either Pushshift.io Reddit or Wikipedia between occupation genderedness and gender mentioning, see Figure~\ref{fig:ablation}. We cannot verify a gender-occupation version of the ``blue banana'' hypothesis, and gender mentions appear to be affected by occupation genderness instead of surprise.

\subsection{Strongest correlations from Pushshift.io Reddit}\label{sec:corpusresults}
Our findings that gender mentioning is correlated with femaleness of occupation are not equally present for all tested corpora: See Figure~\ref{fig:gender_mentioning} and compare Figures~\ref{fig:salientgender}-\ref{fig:salientgenderwiki} in the Appendix. We find that the correlations are medium sized for Pushshift.io Reddit, but very small for Wikipedia and even weaker for \llama Wikipedia. We suspect the difference in correlation strength may be due to dataset size or to domain differences. On size, the Pushshift.io Reddit dataset is several times larger than the Wikipedia datasets, meaning there may be more signal to detect in Pushshift.io Reddit. On text domain, Pushshift.io Reddit has little centralized moderation and is very informal, while Wikipedia has clear style guidelines and an encyclopedic style. For example, Wikipedia policy describes ``neutral point of view'', or a prohibition on subjectivity, which may generally decrease the prevalence of adjectives thought to be evaluative or unnecessary.  

\subsection{Female-coded vs female-dominated}\label{sec:coded}

Above, we compared ground truth occupation gender breakdown (from the US labor statistics) against gender mentioning. This presumes that the ground truth statistics accurately represent the way humans think about occupations and gender, although this may not always be the case. For example, \citet{kotek-etal-2023-gender} reported that people prefer stereotypical gender association with occupations over genders that actually match the ground truth.  
As an expression of this mismatch, people may presume, for example, that the typical `CEO' is a man in our society, even when the occupation is not man-dominated according to our empirical statistics. 
To determine whether peoples' impression of an occupation's gender (and genderedness) differs from the empirical gender (and genderedness), we utilize cosine similarity as a measure of similarity. Following \citet{bolukbasi-etal-2016-man} and \citet{caliskan-etal-2017-semantics}, we deem occupations female- or male-\textbf{coded} by computing cosine distance of word embeddings of an occupation and (fe)male attribute words (see more details in Appendix \ref{sec:femcodewomendom}).
Depending on the distance between an occupation and an attribute word, we can resplit our occupation list into two groups (male- and female- coded), and compute the average mutual information for both.
For both Pushshift.io Reddit and Wikipedia, we find a significant group difference between the mutual information of female- and male- coded occupations.
The mutual information for the female coded occupations is $0.0187 \pm 0.0024$ bits and $0.0161 \pm 0.0012$ bits, respectively. 
In contrast, the mutual information for the male coded occupations is $0.0026 \pm 0.00003$ bits and $0.0037 \pm 0.00013$ bits.
In both cases, the female- vs male- coded groups are about an order of magnitude apart, indicating further evidence that mentioning is more correlated with perceived female occupations.

%Upon analysis, we observe that LL2-Wikipedia uses fewer gendered adjectives to modify job nouns. Specifically, the proportion of jobs referenced with gendered adjectives dropped from 0.54\% to 0.4\%. As gender mentions in this context are often superfluous, this could be seen as possibly desirable behavior. We also observe that, when gender is mentioned, there is even more skew towards using female adjectives over male adjectives in LL2-Wikipedia as compared to the original Wikipedia. This preference was evident in the increase in the proportion of gender-modified job nouns that were modified with female adjectives, rising from an average of 73\% to 83\%. You may find the full result in Appendix \ref{appendix:llamatable}.

%For this, we first look at a possibly linear correlation between occupation genderness and gender mentioning, see Figure~ \ref{fig:gender_mentioing}. Here we find no correlation, hence we consider the null hypothesis to be rejected.

\subsection{What is discussed when people mention gender?}\label{sec:sentiment}

One of our main findings is that occupations associated with women are more likely to be modified with adjectives signaling gender. However, these findings only point to the existence of an effect, they do not tell us, qualitatively, whether the gender mentions occur because of semantic trends present in the comments. 

During qualitative exploration of the Pushshift.io Reddit dataset, we noticed that the vast majority of samples containing woman-dominated occupations (for ``female/male nurse'', for example) included derogatory, and/or sexually explicit content. Others examples discussed gender balance (or lack thereof) in particular occupations, or the experience of people holding non-gender typical roles. Another trend for sentences containing a woman-dominated occupation modified with ``male'' is that these were often negative sentiment. In a smaller number of examples, we also noticed discussions where the gender of a caregiver might be relevant (people expressing that they would prefer a gender matched nurse to care for them during a hospital stay). 

To adduce a bit more quantitative description of these anecdotal observations, we sampled 80 comment threads that mentioned occupations and contained an \texttt{amod} dependency with a gender adjective (20 of each types) to better understand what was being discussed when gender was mentioned. We performed three quantitative measurements: first, we measured comment sentiment, using the Stanza sentiment analysis tool \citep{qi2020stanza}. Next, we handcoded the samples for whether they contain offensive or inappropriate content, and/or whether the topic of discussion is gender balance in occupations. As anticipated, we find that sentiment is higher ($0.6$) when the occupation is man-dominated and modified with ``male'' than for the other three types ($0.4$ for ``female carpenter'' and ``male nuse'', and $0.3$ for ``female nurse''). See Table~\ref{tab:80comments} for more summary statistics for percentage offensive and percentage discussing gender, respectively.

We also ran a logistic regression on offensiveness and occupation gender discussions, to visualize how strongly they might affect gender mentioning. In Figure~\ref{fig:salientgender}, we report that more offensive language was present in discussions pertaining to woman-coded occupations like ``nurse'' than in man-coded ones like ``carpenter''. Gender-related conversations also highlight non-typical gender adjectives modifying occupations (e.g. ``female carpenter'' and ``male nurse''). In sum, these exploratory results indicate that our main findings likely derive from  offensive or negative sentiment conversations about women-coded occupations, or general conversations about gender-balance in occupations. 
\section{Related Work}

\paragraph{Information Theory for Corpus Analysis.} Increasingly, corpus analysis works have relied on information theoretic tools, particularly those pertaining to the lexical semantics of grammatical gender and related morphological specifications \citep{liu-etal-2019-idiosyncrasies,mccarthy-etal-2020-measuring,williams-etal-2019-quantifying,williams-etal-2020-predicting,rathi-etal-2021-information, williams-etal-2021-relationships, chen-etal-2022-investigating, stanczak-etal-2023-grammatical}. 

\paragraph{Gender in Occupations.}

Occupations are interesting for studying gender in NLP. Early work on gender bias in word embeddings  \citep{bolukbasi-etal-2016-man,caliskan-etal-2017-semantics} spawned a wealth of work on social bias and occupations in sentiment analysis \citep{bhaskaran-bhallamudi-2019-good}, coreference resolution \citep{zhao-etal-2018-gender, rudinger-etal-2018-gender}, probing for gender bias \citep{touileb-etal-2022-occupational}, and multilingual applications \citep{stanovsky-etal-2019-evaluating, prates-etal-2020-assessing, troles-schmid-2021-extending, corral-saralegi-2022-gender}.

\section{Conclusion}
We perform a corpus statistical analysis of Wikipedia and Pushshift.io Reddit, and find that there is a relationship between gender mentioning and occupation genderedness using information theoretic techniques. Unlike in other contexts, we find no evidence that gender mentioning is correlated with gender surprise. Instead, we find evidence that gender is more likely to be mentioned for woman-dominated occupations than man-dominated ones.

%\section*{Acknowledgments}

\section{Limitations}

While we have relied pretty heavily on Wikipedia, Pushshift.io Reddit, and the U.S. Labor Statistics as the basis of many of our quantities, other options may be possible. Some issues include: Wikipedia and the Labor statistics do not come from the same joint distribution, but are rather proxies with different bias problems. For instance, English Wikipedia has a fame bias (which could mean our estimates for less lucrative or impressive occupations are noisier), and the labor statistics has a location bias (being US specific, but not English specific). Moreover, there is potentially some temporal discrepancy between these two data sources, in that the Labor Statistics numbers are specific to a recent year (2023), but Wikipedia and Pushshift.io Reddit have been collectively edited and added to over decades.

As with all corpus analyses, our conclusions are limited by the available resources. The occupations we consider are restricted to conventionalized job titles, and therefore we may not be able observe the full range of gender mentioning phenomena. However, what corpus analyses lack in flexibility, they make up for in scale: this work is able to draw conclusions from a large corpus of existing text produced by English speakers, which can give us insights into occupation gender typicality.

\section{Acknowledgments}
We thank Mark Tygert for helpful discussions about our statistical methodology. We also thank our anonymous reviewers for suggesting a few interesting experiments, which we have incorporated into the paper in Section~\ref{sec:sentiment} and the Appendix. 

\bibliography{custom, anthology}

\begin{thebibliography}{44}
\expandafter\ifx\csname natexlab\endcsname\relax\def\natexlab#1{#1}\fi

\bibitem[{Ackerman(2019)}]{ackerman2019syntactic}
Lauren~M Ackerman. 2019.
\newblock Syntactic and cognitive issues in investigating gendered coreference.
\newblock \emph{Glossa}.

\bibitem[{Attardi(2015)}]{Wikiextractor2015}
Giusepppe Attardi. 2015.
\newblock Wikiextractor.
\newblock \url{https://github.com/attardi/wikiextractor}.

\bibitem[{Bailey et~al.(2019)Bailey, LaFrance, and Dovidio}]{bailey2019man}
April~H Bailey, Marianne LaFrance, and John~F Dovidio. 2019.
\newblock Is man the measure of all things? a social cognitive account of androcentrism.
\newblock \emph{Personality and Social Psychology Review}, 23(4):307--331.

\bibitem[{Bailey et~al.(2020)Bailey, LaFrance, and Dovidio}]{bailey2020implicit}
April~H Bailey, Marianne LaFrance, and John~F Dovidio. 2020.
\newblock Implicit androcentrism: Men are human, women are gendered.
\newblock \emph{Journal of Experimental Social Psychology}, 89:103980.

\bibitem[{Bailey et~al.(2022)Bailey, Williams, and Cimpian}]{bailey2022based}
April~H Bailey, Adina Williams, and Andrei Cimpian. 2022.
\newblock Based on billions of words on the internet, people= men.
\newblock \emph{Science Advances}, 8(13):eabm2463.

\bibitem[{Bailey et~al.(2024)Bailey, Williams, Poddar, and Cimpian}]{bailey2024intersectional}
April~H Bailey, Adina Williams, Aashna Poddar, and Andrei Cimpian. 2024.
\newblock Intersectional male-centric and white-centric biases in collective concepts.

\bibitem[{Bartl et~al.(2020)Bartl, Nissim, and Gatt}]{bartl-etal-2020-unmasking}
Marion Bartl, Malvina Nissim, and Albert Gatt. 2020.
\newblock \href {https://aclanthology.org/2020.gebnlp-1.1} {Unmasking contextual stereotypes: Measuring and mitigating {BERT}{'}s gender bias}.
\newblock In \emph{Proceedings of the Second Workshop on Gender Bias in Natural Language Processing}, pages 1--16, Barcelona, Spain (Online). Association for Computational Linguistics.

\bibitem[{Baumgartner et~al.(2020)Baumgartner, Zannettou, Keegan, Squire, and Blackburn}]{Baumgartner_Zannettou_Keegan_Squire_Blackburn_2020}
Jason Baumgartner, Savvas Zannettou, Brian Keegan, Megan Squire, and Jeremy Blackburn. 2020.
\newblock \href {https://doi.org/10.1609/icwsm.v14i1.7347} {The pushshift reddit dataset}.
\newblock \emph{Proceedings of the International AAAI Conference on Web and Social Media}, 14(1):830--839.

\bibitem[{Bem(1993)}]{bem1993lenses}
Sandra~L. Bem. 1993.
\newblock \emph{The lenses of gender: Transforming the debate on sexual inequality}.
\newblock Yale University Press.

\bibitem[{Bhaskaran and Bhallamudi(2019)}]{bhaskaran-bhallamudi-2019-good}
Jayadev Bhaskaran and Isha Bhallamudi. 2019.
\newblock \href {https://doi.org/10.18653/v1/W19-3809} {Good secretaries, bad truck drivers? occupational gender stereotypes in sentiment analysis}.
\newblock In \emph{Proceedings of the First Workshop on Gender Bias in Natural Language Processing}, pages 62--68, Florence, Italy. Association for Computational Linguistics.

\bibitem[{Bolukbasi et~al.(2016)Bolukbasi, Chang, Zou, Saligrama, and Kalai}]{bolukbasi-etal-2016-man}
Tolga Bolukbasi, Kai-Wei Chang, James~Y Zou, Venkatesh Saligrama, and Adam~T Kalai. 2016.
\newblock Man is to computer programmer as woman is to homemaker? debiasing word embeddings.
\newblock \emph{Advances in neural information processing systems}, 29.

\bibitem[{Caliskan et~al.(2017)Caliskan, Bryson, and Narayanan}]{caliskan-etal-2017-semantics}
Aylin Caliskan, Joanna~J Bryson, and Arvind Narayanan. 2017.
\newblock Semantics derived automatically from language corpora contain human-like biases.
\newblock \emph{Science}, 356(6334):183--186.

\bibitem[{Chen and Manning(2014)}]{chen-manning-2014-fast}
Danqi Chen and Christopher Manning. 2014.
\newblock \href {https://doi.org/10.3115/v1/D14-1082} {A fast and accurate dependency parser using neural networks}.
\newblock In \emph{Proceedings of the 2014 Conference on Empirical Methods in Natural Language Processing ({EMNLP})}, pages 740--750, Doha, Qatar. Association for Computational Linguistics.

\bibitem[{Chen et~al.(2022)Chen, Futrell, and Mahowald}]{chen-etal-2022-investigating}
Sihan Chen, Richard Futrell, and Kyle Mahowald. 2022.
\newblock \href {https://doi.org/10.18653/v1/2022.sigtyp-1.12} {Investigating information-theoretic properties of the typology of spatial demonstratives}.
\newblock In \emph{Proceedings of the 4th Workshop on Research in Computational Linguistic Typology and Multilingual NLP}, pages 94--95, Seattle, Washington. Association for Computational Linguistics.

\bibitem[{Corral and Saralegi(2022)}]{corral-saralegi-2022-gender}
Ander Corral and Xabier Saralegi. 2022.
\newblock \href {https://aclanthology.org/2022.wmt-1.10} {Gender bias mitigation for {NMT} involving genderless languages}.
\newblock In \emph{Proceedings of the Seventh Conference on Machine Translation (WMT)}, pages 165--176, Abu Dhabi, United Arab Emirates (Hybrid). Association for Computational Linguistics.

\bibitem[{Degen et~al.(2020)Degen, Hawkins, Graf, Kreiss, and Goodman}]{Degen2020redundancy}
Judith Degen, Robert~D Hawkins, Caroline Graf, Elisa Kreiss, and Noah~D Goodman. 2020.
\newblock When redundancy is useful: A bayesian approach to “overinformative” referring expressions.
\newblock \emph{Psychological Review}, 127(4):591.

\bibitem[{Falenska and {\c{C}}etino{\u{g}}lu(2021)}]{falenska-cetinoglu-2021-assessing}
Agnieszka Falenska and {\"O}zlem {\c{C}}etino{\u{g}}lu. 2021.
\newblock \href {https://doi.org/10.18653/v1/2021.gebnlp-1.9} {Assessing gender bias in {W}ikipedia: Inequalities in article titles}.
\newblock In \emph{Proceedings of the 3rd Workshop on Gender Bias in Natural Language Processing}, pages 75--85, Online. Association for Computational Linguistics.

\bibitem[{Gilman(1911)}]{gilman1911man}
Charlotte~Perkins Gilman. 1911.
\newblock \emph{The man-made world}.
\newblock Charlotte Company.

\bibitem[{Gonz{\'a}lez et~al.(2020)Gonz{\'a}lez, Barrett, Hvingelby, Webster, and S{\o}gaard}]{gonzalez-etal-2020-type}
Ana~Valeria Gonz{\'a}lez, Maria Barrett, Rasmus Hvingelby, Kellie Webster, and Anders S{\o}gaard. 2020.
\newblock \href {https://doi.org/10.18653/v1/2020.emnlp-main.209} {Type {B} reflexivization as an unambiguous testbed for multilingual multi-task gender bias}.
\newblock In \emph{Proceedings of the 2020 Conference on Empirical Methods in Natural Language Processing (EMNLP)}, pages 2637--2648, Online. Association for Computational Linguistics.

\bibitem[{Hamilton(1991)}]{hamilton1991masculine}
Mykol~C. Hamilton. 1991.
\newblock Masculine bias in the attribution of personhood: People= male, male= people.
\newblock \emph{Psychology of Women Quarterly}, 15(3):393--402.

\bibitem[{Hoyle et~al.(2019)Hoyle, Wolf-Sonkin, Wallach, Augenstein, and Cotterell}]{hoyle-etal-2019-unsupervised}
Alexander~Miserlis Hoyle, Lawrence Wolf-Sonkin, Hanna Wallach, Isabelle Augenstein, and Ryan Cotterell. 2019.
\newblock \href {https://doi.org/10.18653/v1/P19-1167} {Unsupervised discovery of gendered language through latent-variable modeling}.
\newblock In \emph{Proceedings of the 57th Annual Meeting of the Association for Computational Linguistics}, pages 1706--1716, Florence, Italy. Association for Computational Linguistics.

\bibitem[{Humeau et~al.(2020)Humeau, Shuster, Lachaux, and Weston}]{humeau2020polyencoders}
Samuel Humeau, Kurt Shuster, Marie-Anne Lachaux, and Jason Weston. 2020.
\newblock \href {http://arxiv.org/abs/1905.01969} {Poly-encoders: Transformer architectures and pre-training strategies for fast and accurate multi-sentence scoring}.

\bibitem[{Kotek et~al.(2023)Kotek, Dockum, and Sun}]{kotek-etal-2023-gender}
Hadas Kotek, Rikker Dockum, and David Sun. 2023.
\newblock Gender bias and stereotypes in large language models.
\newblock In \emph{Proceedings of The ACM Collective Intelligence Conference}, pages 12--24.

\bibitem[{Liu et~al.(2019)Liu, Mei, Williams, and Cotterell}]{liu-etal-2019-idiosyncrasies}
Shijia Liu, Hongyuan Mei, Adina Williams, and Ryan Cotterell. 2019.
\newblock \href {https://doi.org/10.18653/v1/N19-1415} {On the idiosyncrasies of the {M}andarin {C}hinese classifier system}.
\newblock In \emph{Proceedings of the 2019 Conference of the North {A}merican Chapter of the Association for Computational Linguistics: Human Language Technologies, Volume 1 (Long and Short Papers)}, pages 4100--4106, Minneapolis, Minnesota. Association for Computational Linguistics.

\bibitem[{McCarthy et~al.(2020)McCarthy, Williams, Liu, Yarowsky, and Cotterell}]{mccarthy-etal-2020-measuring}
Arya~D. McCarthy, Adina Williams, Shijia Liu, David Yarowsky, and Ryan Cotterell. 2020.
\newblock \href {https://doi.org/10.18653/v1/2020.emnlp-main.456} {Measuring the similarity of grammatical gender systems by comparing partitions}.
\newblock In \emph{Proceedings of the 2020 Conference on Empirical Methods in Natural Language Processing (EMNLP)}, pages 5664--5675, Online. Association for Computational Linguistics.

\bibitem[{Merritt and Kok(1995)}]{merritt1995attribution}
Rebecca~Davis Merritt and Cynthia~J. Kok. 1995.
\newblock Attribution of gender to a gender-unspecified individual: An evaluation of the people= male hypothesis.
\newblock \emph{Sex Roles}, 33:145--157.

\bibitem[{Polyanskiy and Wu(2014)}]{polyanskiy2015lecture}
Y~Polyanskiy and Y~Wu. 2014.
\newblock Lecture notes on information theory.

\bibitem[{Prates et~al.(2020)Prates, Avelar, and Lamb}]{prates-etal-2020-assessing}
Marcelo~OR Prates, Pedro~H Avelar, and Lu{\'\i}s~C Lamb. 2020.
\newblock Assessing gender bias in machine translation: a case study with google translate.
\newblock \emph{Neural Computing and Applications}, 32:6363--6381.

\bibitem[{Qi et~al.(2020)Qi, Zhang, Zhang, Bolton, and Manning}]{qi2020stanza}
Peng Qi, Yuhao Zhang, Yuhui Zhang, Jason Bolton, and Christopher~D. Manning. 2020.
\newblock \href {http://arxiv.org/abs/2003.07082} {Stanza: A python natural language processing toolkit for many human languages}.

\bibitem[{Rathi et~al.(2021)Rathi, Hahn, and Futrell}]{rathi-etal-2021-information}
Neil Rathi, Michael Hahn, and Richard Futrell. 2021.
\newblock \href {https://doi.org/10.18653/v1/2021.emnlp-main.793} {An information-theoretic characterization of morphological fusion}.
\newblock In \emph{Proceedings of the 2021 Conference on Empirical Methods in Natural Language Processing}, pages 10115--10120, Online and Punta Cana, Dominican Republic. Association for Computational Linguistics.

\bibitem[{Roller et~al.(2020)Roller, Dinan, Goyal, Ju, Williamson, Liu, Xu, Ott, Shuster, Smith, Boureau, and Weston}]{roller2020recipes}
Stephen Roller, Emily Dinan, Naman Goyal, Da~Ju, Mary Williamson, Yinhan Liu, Jing Xu, Myle Ott, Kurt Shuster, Eric~M. Smith, Y-Lan Boureau, and Jason Weston. 2020.
\newblock \href {http://arxiv.org/abs/2004.13637} {Recipes for building an open-domain chatbot}.

\bibitem[{Rudinger et~al.(2018)Rudinger, Naradowsky, Leonard, and Van~Durme}]{rudinger-etal-2018-gender}
Rachel Rudinger, Jason Naradowsky, Brian Leonard, and Benjamin Van~Durme. 2018.
\newblock \href {https://doi.org/10.18653/v1/N18-2002} {Gender bias in coreference resolution}.
\newblock In \emph{Proceedings of the 2018 Conference of the North {A}merican Chapter of the Association for Computational Linguistics: Human Language Technologies, Volume 2 (Short Papers)}, pages 8--14, New Orleans, Louisiana. Association for Computational Linguistics.

\bibitem[{Schmahl et~al.(2020)Schmahl, Viering, Makrodimitris, Naseri~Jahfari, Tax, and Loog}]{schmahl-etal-2020-wikipedia}
Katja~Geertruida Schmahl, Tom~Julian Viering, Stavros Makrodimitris, Arman Naseri~Jahfari, David Tax, and Marco Loog. 2020.
\newblock \href {https://doi.org/10.18653/v1/2020.nlpcss-1.11} {Is {W}ikipedia succeeding in reducing gender bias? assessing changes in gender bias in {W}ikipedia using word embeddings}.
\newblock In \emph{Proceedings of the Fourth Workshop on Natural Language Processing and Computational Social Science}, pages 94--103, Online. Association for Computational Linguistics.

\bibitem[{Sta{\'n}czak et~al.(2023)Sta{\'n}czak, Du, Williams, Augenstein, and Cotterell}]{stanczak-etal-2023-grammatical}
Karolina Sta{\'n}czak, Kevin Du, Adina Williams, Isabelle Augenstein, and Ryan Cotterell. 2023.
\newblock Grammatical gender's influence on distributional semantics: A causal perspective.
\newblock \emph{arXiv preprint arXiv:2311.18567}.

\bibitem[{Stanovsky et~al.(2019)Stanovsky, Smith, and Zettlemoyer}]{stanovsky-etal-2019-evaluating}
Gabriel Stanovsky, Noah~A. Smith, and Luke Zettlemoyer. 2019.
\newblock \href {https://doi.org/10.18653/v1/P19-1164} {Evaluating gender bias in machine translation}.
\newblock In \emph{Proceedings of the 57th Annual Meeting of the Association for Computational Linguistics}, pages 1679--1684, Florence, Italy. Association for Computational Linguistics.

\bibitem[{Touileb et~al.(2022)Touileb, {\O}vrelid, and Velldal}]{touileb-etal-2022-occupational}
Samia Touileb, Lilja {\O}vrelid, and Erik Velldal. 2022.
\newblock \href {https://doi.org/10.18653/v1/2022.gebnlp-1.21} {Occupational biases in {N}orwegian and multilingual language models}.
\newblock In \emph{Proceedings of the 4th Workshop on Gender Bias in Natural Language Processing (GeBNLP)}, pages 200--211, Seattle, Washington. Association for Computational Linguistics.

\bibitem[{Touvron et~al.(2023)Touvron, Martin, Stone, Albert, Almahairi, Babaei, Bashlykov, Batra, Bhargava, Bhosale, Bikel, Blecher, Canton{-}Ferrer, Chen, Cucurull, Esiobu, Fernandes, Fu, Fu, Fuller, Gao, Goswami, Goyal, Hartshorn, Hosseini, Hou, Inan, Kardas, Kerkez, Khabsa, Kloumann, Korenev, Koura, Lachaux, Lavril, Lee, Liskovich, Lu, Mao, Martinet, Mihaylov, Mishra, Molybog, Nie, Poulton, Reizenstein, Rungta, Saladi, Schelten, Silva, Smith, Subramanian, Tan, Tang, Taylor, Williams, Kuan, Xu, Yan, Zarov, Zhang, Fan, Kambadur, Narang, Rodriguez, Stojnic, Edunov, and Scialom}]{2023-llama2}
Hugo Touvron, Louis Martin, Kevin Stone, Peter Albert, Amjad Almahairi, Yasmine Babaei, Nikolay Bashlykov, Soumya Batra, Prajjwal Bhargava, Shruti Bhosale, Dan Bikel, Lukas Blecher, Cristian Canton{-}Ferrer, Moya Chen, Guillem Cucurull, David Esiobu, Jude Fernandes, Jeremy Fu, Wenyin Fu, Brian Fuller, Cynthia Gao, Vedanuj Goswami, Naman Goyal, Anthony Hartshorn, Saghar Hosseini, Rui Hou, Hakan Inan, Marcin Kardas, Viktor Kerkez, Madian Khabsa, Isabel Kloumann, Artem Korenev, Punit~Singh Koura, Marie{-}Anne Lachaux, Thibaut Lavril, Jenya Lee, Diana Liskovich, Yinghai Lu, Yuning Mao, Xavier Martinet, Todor Mihaylov, Pushkar Mishra, Igor Molybog, Yixin Nie, Andrew Poulton, Jeremy Reizenstein, Rashi Rungta, Kalyan Saladi, Alan Schelten, Ruan Silva, Eric~Michael Smith, Ranjan Subramanian, Xiaoqing~Ellen Tan, Binh Tang, Ross Taylor, Adina Williams, Jian~Xiang Kuan, Puxin Xu, Zheng Yan, Iliyan Zarov, Yuchen Zhang, Angela Fan, Melanie Kambadur, Sharan Narang, Aur{\'{e}}lien Rodriguez, Robert Stojnic, Sergey Edunov,
  and Thomas Scialom. 2023.
\newblock \href {https://doi.org/10.48550/ARXIV.2307.09288} {Llama 2: Open foundation and fine-tuned chat models}.
\newblock \emph{CoRR}, abs/2307.09288.

\bibitem[{Troles and Schmid(2021)}]{troles-schmid-2021-extending}
Jonas-Dario Troles and Ute Schmid. 2021.
\newblock \href {https://aclanthology.org/2021.wmt-1.61} {Extending challenge sets to uncover gender bias in machine translation: Impact of stereotypical verbs and adjectives}.
\newblock In \emph{Proceedings of the Sixth Conference on Machine Translation}, pages 531--541, Online. Association for Computational Linguistics.

\bibitem[{Wagner et~al.(2015)Wagner, Garcia, Jadidi, and Strohmaier}]{wagner2015its}
Claudia Wagner, David Garcia, Mohsen Jadidi, and Markus Strohmaier. 2015.
\newblock It's a man's wikipedia? assessing gender inequality in an online encyclopedia.
\newblock In \emph{Proceedings of the international AAAI conference on web and social media}, volume~9, pages 454--463.

\bibitem[{Williams et~al.(2019)Williams, Blasi, Wolf-Sonkin, Wallach, and Cotterell}]{williams-etal-2019-quantifying}
Adina Williams, Damian Blasi, Lawrence Wolf-Sonkin, Hanna Wallach, and Ryan Cotterell. 2019.
\newblock \href {https://doi.org/10.18653/v1/D19-1577} {Quantifying the semantic core of gender systems}.
\newblock In \emph{Proceedings of the 2019 Conference on Empirical Methods in Natural Language Processing and the 9th International Joint Conference on Natural Language Processing (EMNLP-IJCNLP)}, pages 5734--5739, Hong Kong, China. Association for Computational Linguistics.

\bibitem[{Williams et~al.(2021)Williams, Cotterell, Wolf-Sonkin, Blasi, and Wallach}]{williams-etal-2021-relationships}
Adina Williams, Ryan Cotterell, Lawrence Wolf-Sonkin, Dami{\'a}n Blasi, and Hanna Wallach. 2021.
\newblock \href {https://doi.org/10.1162/tacl_a_00355} {On the relationships between the grammatical genders of inanimate nouns and their co-occurring adjectives and verbs}.
\newblock \emph{Transactions of the Association for Computational Linguistics}, 9:139--159.

\bibitem[{Williams et~al.(2020)Williams, Pimentel, Blix, McCarthy, Chodroff, and Cotterell}]{williams-etal-2020-predicting}
Adina Williams, Tiago Pimentel, Hagen Blix, Arya~D. McCarthy, Eleanor Chodroff, and Ryan Cotterell. 2020.
\newblock \href {https://doi.org/10.18653/v1/2020.acl-main.597} {Predicting declension class from form and meaning}.
\newblock In \emph{Proceedings of the 58th Annual Meeting of the Association for Computational Linguistics}, pages 6682--6695, Online. Association for Computational Linguistics.

\bibitem[{Wilson and Meyer(2021)}]{wilson2021nonbinary}
Bianca~D.M. Wilson and Ilan~H. Meyer. 2021.
\newblock Nonbinary {LGBTQ} adults in the {U}nited {S}tates.
\newblock \url{https://williamsinstitute.law.ucla.edu/publications/nonbinary-lgbtq-adults-us/}.

\bibitem[{Zhao et~al.(2018)Zhao, Wang, Yatskar, Ordonez, and Chang}]{zhao-etal-2018-gender}
Jieyu Zhao, Tianlu Wang, Mark Yatskar, Vicente Ordonez, and Kai-Wei Chang. 2018.
\newblock \href {https://doi.org/10.18653/v1/N18-2003} {Gender bias in coreference resolution: Evaluation and debiasing methods}.
\newblock In \emph{Proceedings of the 2018 Conference of the North {A}merican Chapter of the Association for Computational Linguistics: Human Language Technologies, Volume 2 (Short Papers)}, pages 15--20, New Orleans, Louisiana. Association for Computational Linguistics.

\end{thebibliography}

\clearpage
\appendix
\section{Notation} \label{appdendix:notation}

Throughout this work, we follow \citet{polyanskiy2015lecture} for our notation.  We denote a probability space as $(\Omega, \mathcal{F}, \mathbb{P})$. We define random variables (r.v.) as $X: \Omega \to \mathcal{X}$. We denote an r.v. as a capital letter, $X$, while their realizations are denoted by a lower case letter, $x$. 
The distribution of $X$ is represented as $P_X$, which is a probability measure on the alphabet $\mathcal{X}$. We use $p_X$ for probability mass function of $P_X$, we may often drop the subscript if the context is unambiguous.
%The probability density (or mass) function of $P_X$ is denoted as $p_X$, and we may drop the subscript when there is no ambiguity. 
We use $P_{X|Y}$ for a conditional distribution, which can be thought of as a collection of probability measures on $\mathcal{X}$, one $P_{X|Y=y}$ for each value of $y$. 
Expectations are denoted either as $\mathbb{E}[X]$, or $\mathbb{E}_{x\sim P_X}[x]$, and similarly, $\mathbb{E}[f(X)]$ can also be denoted by $\mathbb{E}_{x\sim P_X}[f(x)]$.
A concept of particular importance in this study is the entropy of a random variable, which is defined as the expected information content of a random variable $X$; $H(X) = \mathbb{E}_{x\sim P_X}[- \log_2 p(x)]$.
We further will rely on mutual information, denoted as $I(X;Y)$. It quantifies the mutual dependence between two random variables $X$ and $Y$. It is defined as $I(X;Y) = \mathbb{E}_{(x,y)\sim P_{X,Y}}[\log_2 \frac{p_{X,Y}(x,y)}{p_X(x)p_Y(y)}]$ and represents the reduction in uncertainty about $X$ given the knowledge of $Y$, and vice versa. 

\section{List of nouns}
\label{sec:listofnouns}
\subsection{Occupation}
\label{sec:listofoccupation}

accountant
analyst
assistant
attendant
auditor
baker
carpenter
cashier
ceo
chief
cleaner
clerk
constructor
cook
counselor
designer
developer
driver
editor
farmer
guard
hairdresser
housekeeper
janitor
laborer
lawyer
librarian
manager
%mechanician
mover
nurse
physician
receptionist
salesperson
secretary
sheriff
supervisor
teacher
writer

\section{Prompt for \llama Wikipedia}
\label{sec:promptllamawikipedia}
Generate a Wikipedia article on the topic of \{title\}. 

Use the following first paragraph from the original Wikipedia article as a starting point:

\{first 256 words\}

Now, expand upon the provided paragraph by providing additional details, 
historical context, notable events, key figures, and any relevant subtopics. 
Aim for a well-structured and informative Wikipedia style article with a minimum length of 700 words. Ensure that the content is factually accurate, well-written, and on Wikipedia writing style.

\section{BLEU score between Wikipedia and \llama Wikipedia}
\label{appendix:bleuscore}
\begin{table}[!h]
\centering
\begin{tabular}{lccc}
\toprule
 & BLEU-2 & BLEU-3 & BLEU-4 \\
\midrule
Median & 0.13 & 0.11 & 0.09 \\
Mean& 0.16 & 0.14& 0.12 \\
95 percentile & 0.39 & 0.35 & 0.33\\
\bottomrule
\end{tabular}
\caption{The table shows the mean, median, and 95th percentile of BLEU scores between original and \llama Wikipedia articles. The observed low token overlap signifies substantial linguistic diversity, thereby establishing \llama Wikipedia as a valuable dataset for analysis.}
\label{tab:bleuscorebetweenwikipedias}
\end{table}

\clearpage
\onecolumn
\section{Occupation mentions}
\subsection{Pushshift.io Reddit}\label{appendix:reddittable}
\begin{table}[!ht]
    \centering
    \begin{tabular}{lccccccc}
    \toprule
        Occupation & Total & Male & Female & Total gendered & Gendered \% & Male \% & Female \% \\ 
        \hline
        accountant & 429663 & 581 & 430 & 1011 & 0.24\% & 57.47\% & 42.53\% \\ 
        analyst & 661383 & 415 & 428 & 843 & 0.13\% & 49.23\% & 50.77\% \\ 
        assistant & 1090414 & 1878 & 2764 & 4642 & 0.43\% & 40.46\% & 59.54\% \\ 
        % attendant & 288544 & 2013 & 1510 & 3523 & 1.22\% & 57.14\% & 42.86\% \\ 
        auditor & 91014 & 52 & 51 & 103 & 0.11\% & 50.49\% & 49.51\% \\ 
        baker & 569508 & 360 & 206 & 566 & 0.10\% & 63.60\% & 36.40\% \\ 
        carpenter & 278880 & 454 & 315 & 769 & 0.28\% & 59.04\% & 40.96\% \\ 
        cashier & 708872 & 1725 & 2789 & 4514 & 0.64\% & 38.21\% & 61.79\% \\ 
        ceo & 1395508 & 4397 & 9846 & 14243 & 1.02\% & 30.87\% & 69.13\% \\ 
        chief & 1889533 & 1671 & 1168 & 2839 & 0.15\% & 58.86\% & 41.14\% \\ 
        cleaner & 781771 & 664 & 527 & 1191 & 0.15\% & 55.75\% & 44.25\% \\ 
        clerk & 406230 & 771 & 1357 & 2128 & 0.52\% & 36.23\% & 63.77\% \\ 
        constructor & 117540 & 83 & 123 & 206 & 0.18\% & 40.29\% & 59.71\% \\ 
        cook & 6087570 & 9972 & 7298 & 17270 & 0.28\% & 57.74\% & 42.26\% \\ 
        counselor & 870469 & 1481 & 1732 & 3213 & 0.37\% & 46.09\% & 53.91\% \\ 
        designer & 1573543 & 1968 & 2019 & 3987 & 0.25\% & 49.36\% & 50.64\% \\ 
        developer & 4871656 & 2996 & 6189 & 9185 & 0.19\% & 32.62\% & 67.38\% \\ 
        driver & 9852935 & 15025 & 31213 & 46238 & 0.47\% & 32.49\% & 67.51\% \\ 
        editor & 1315447 & 1004 & 1072 & 2076 & 0.16\% & 48.36\% & 51.64\% \\ 
        farmer & 1608655 & 1923 & 1059 & 2982 & 0.19\% & 64.49\% & 35.51\% \\ 
        guard & 5748190 & 13506 & 8836 & 22342 & 0.39\% & 60.45\% & 39.55\% \\ 
        hairdresser & 106245 & 752 & 681 & 1433 & 1.35\% & 52.48\% & 47.52\% \\ 
        housekeeper & 55326 & 197 & 283 & 480 & 0.87\% & 41.04\% & 58.96\% \\ 
        janitor & 278991 & 1257 & 582 & 1839 & 0.66\% & 68.35\% & 31.65\% \\ 
        laborer & 150724 & 394 & 256 & 650 & 0.43\% & 60.62\% & 39.38\% \\ 
        lawyer & 4087070 & 5045 & 7446 & 12491 & 0.31\% & 40.39\% & 59.61\% \\ 
        librarian & 236107 & 489 & 478 & 967 & 0.41\% & 50.57\% & 49.43\% \\ 
        manager & 6103567 & 10731 & 10092 & 20823 & 0.34\% & 51.53\% & 48.47\% \\ 
     %   mechanician & 126 & 0 & 0 & 0 & 0.00\% & NA & NA \\ 
        mover & 185850 & 319 & 103 & 422 & 0.23\% & 75.59\% & 24.41\% \\ 
        nurse & 2355458 & 24441 & 11889 & 36330 & 1.54\% & 67.27\% & 32.73\% \\ 
        physician & 523220 & 1243 & 1820 & 3063 & 0.59\% & 40.58\% & 59.42\% \\ 
        receptionist & 139164 & 440 & 544 & 984 & 0.71\% & 44.72\% & 55.28\% \\ 
        salesperson & 81141 & 123 & 266 & 389 & 0.48\% & 31.62\% & 68.38\% \\ 
        secretary & 623470 & 1405 & 2323 & 3728 & 0.60\% & 37.69\% & 62.31\% \\ 
        sheriff & 418668 & 492 & 265 & 757 & 0.18\% & 64.99\% & 35.01\% \\ 
        supervisor & 733711 & 1453 & 2124 & 3577 & 0.49\% & 40.62\% & 59.38\% \\ 
        teacher & 8204500 & 49877 & 43351 & 93228 & 1.14\% & 53.50\% & 46.50\% \\ 
        writer & 3804680 & 10034 & 14947 & 24981 & 0.66\% & 40.17\% & 59.83\% \\ 
        \bottomrule
    \end{tabular}
    \caption{Statistics of occupation mentions and gender-specific mentions in the Pushshift.io Reddit dataset.}
\end{table}
\clearpage
\subsection{Wikipedia}
\begin{table*}[!ht]
    \centering
    \begin{tabular}{lccccccc}
    \toprule
        Occupation & Total & Male & Female & Total gendered & Gendered \% & Male \% & Female \% \\ 
        \hline
        accountant & 18223 & 16 & 32 & 48 & 0.26\% & 33.33\% & 66.67\% \\ 
        analyst & 39200 & 18 & 58 & 76 & 0.19\% & 23.68\% & 76.32\% \\ 
        assistant & 280051 & 146 & 306 & 452 & 0.16\% & 32.30\% & 67.70\% \\ 
        % attendant & 14958 & 98 & 333 & 431 & 2.88\% & 22.74\% & 77.26\% \\ 
        auditor & 15418 & 3 & 25 & 28 & 0.18\% & 10.71\% & 89.29\% \\ 
        baker & 74802 & 21 & 9 & 30 & 0.04\% & 70.00\% & 30.00\% \\ 
        carpenter & 36691 & 28 & 15 & 43 & 0.12\% & 65.12\% & 34.88\% \\ 
        cashier & 3767 & 1 & 13 & 14 & 0.37\% & 7.14\% & 92.86\% \\ 
        ceo & 103697 & 27 & 289 & 316 & 0.30\% & 8.54\% & 91.46\% \\ 
        chief & 571954 & 288 & 480 & 768 & 0.13\% & 37.50\% & 62.50\% \\ 
        cleaner & 7370 & 9 & 22 & 31 & 0.42\% & 29.03\% & 70.97\% \\ 
        clerk & 55621 & 63 & 150 & 213 & 0.38\% & 29.58\% & 70.42\% \\ 
        constructor & 8910 & 2 & 9 & 11 & 0.12\% & 18.18\% & 81.82\% \\ 
        cook & 102970 & 134 & 249 & 383 & 0.37\% & 34.99\% & 65.01\% \\ 
        counselor & 14953 & 12 & 33 & 45 & 0.30\% & 26.67\% & 73.33\% \\ 
        designer & 115606 & 53 & 242 & 295 & 0.26\% & 17.97\% & 82.03\% \\ 
        developer & 80650 & 19 & 45 & 64 & 0.08\% & 29.69\% & 70.31\% \\ 
        driver & 203909 & 291 & 911 & 1202 & 0.59\% & 24.21\% & 75.79\% \\ 
        editor & 273589 & 120 & 843 & 963 & 0.35\% & 12.46\% & 87.54\% \\ 
        farmer & 138372 & 171 & 322 & 493 & 0.36\% & 34.69\% & 65.31\% \\ 
        guard & 275038 & 1037 & 537 & 1574 & 0.57\% & 65.88\% & 34.12\% \\ 
        hairdresser & 3734 & 20 & 22 & 42 & 1.12\% & 47.62\% & 52.38\% \\ 
        housekeeper & 5326 & 18 & 39 & 57 & 1.07\% & 31.58\% & 68.42\% \\ 
        janitor & 3642 & 10 & 7 & 17 & 0.47\% & 58.82\% & 41.18\% \\ 
        laborer & 13725 & 75 & 119 & 194 & 1.41\% & 38.66\% & 61.34\% \\ 
        lawyer & 147598 & 252 & 1507 & 1759 & 1.19\% & 14.33\% & 85.67\% \\ 
        librarian & 22557 & 18 & 106 & 124 & 0.55\% & 14.52\% & 85.48\% \\ 
        manager & 371536 & 205 & 503 & 708 & 0.19\% & 28.95\% & 71.05\% \\ 
        % mechanician & 147 & 0 & 0 & 0 & 0.00\% & NA & NA \\ 
        mover & 5365 & 10 & 1 & 11 & 0.21\% & 90.91\% & 9.09\% \\ 
        nurse & 64244 & 349 & 480 & 829 & 1.29\% & 42.10\% & 57.90\% \\ 
        physician & 92604 & 198 & 940 & 1138 & 1.23\% & 17.40\% & 82.60\% \\ 
        receptionist & 2405 & 2 & 10 & 12 & 0.50\% & 16.67\% & 83.33\% \\ 
        salesperson & 823 & 0 & 7 & 7 & 0.85\% & 0.00\% & 100.00\% \\ 
        secretary & 343292 & 124 & 552 & 676 & 0.20\% & 18.34\% & 81.66\% \\ 
        sheriff & 55098 & 82 & 62 & 144 & 0.26\% & 56.94\% & 43.06\% \\ 
        supervisor & 32882 & 38 & 61 & 99 & 0.30\% & 38.38\% & 61.62\% \\ 
        teacher & 305614 & 616 & 1404 & 2020 & 0.66\% & 30.50\% & 69.50\% \\ 
        writer & 440711 & 717 & 3770 & 4487 & 1.02\% & 15.98\% & 84.02\% \\ 
        \bottomrule
    \end{tabular}
    \caption{Statistics of occupation mentions and gender-specific mentions in the Wikipedia dataset.}
\end{table*}
\clearpage
\subsection{\llama Wikipedia}
\label{appendix:llamatable}

\begin{table}[!ht]
    \centering
   %\captionsetup{width=0.9\paperwidth}
    \begin{tabular}{lccccccc}
    \toprule
        Occupation & Total & Male & Female & Total gendered & Gendered \% & Male \% & Female \% \\ 
        \hline
        accountant & 20919 & 12 & 40 & 52 & 0.25\% & 23.08\% & 76.92\% \\ 
        analyst & 61961 & 6 & 38 & 44 & 0.07\% & 13.64\% & 86.36\% \\ 
        assistant & 309714 & 77 & 159 & 236 & 0.08\% & 32.63\% & 67.37\% \\ 
        % attendant & 9532 & 39 & 184 & 223 & 2.34\% & 17.49\% & 82.51\% \\ 
        auditor & 13948 & 0 & 15 & 15 & 0.11\% & 0.00\% & 100.00\% \\ 
        baker & 107206 & 13 & 5 & 18 & 0.02\% & 72.22\% & 27.78\% \\ 
        carpenter & 56454 & 19 & 9 & 28 & 0.05\% & 67.86\% & 32.14\% \\ 
        cashier & 2470 & 1 & 6 & 7 & 0.28\% & 14.29\% & 85.71\% \\ 
        ceo & 230598 & 19 & 428 & 447 & 0.19\% & 4.25\% & 95.75\% \\ 
        chief & 685222 & 226 & 658 & 884 & 0.13\% & 25.57\% & 74.43\% \\ 
        cleaner & 7766 & 2 & 8 & 10 & 0.13\% & 20.00\% & 80.00\% \\ 
        clerk & 59117 & 26 & 90 & 116 & 0.20\% & 22.41\% & 77.59\% \\ 
        constructor & 7357 & 1 & 2 & 3 & 0.04\% & 33.33\% & 66.67\% \\ 
        cook & 148143 & 43 & 113 & 156 & 0.11\% & 27.56\% & 72.44\% \\ 
        counselor & 22234 & 9 & 23 & 32 & 0.14\% & 28.13\% & 71.88\% \\ 
        designer & 295526 & 42 & 446 & 488 & 0.17\% & 8.61\% & 91.39\% \\ 
        developer & 183697 & 21 & 80 & 101 & 0.05\% & 20.79\% & 79.21\% \\ 
        driver & 362546 & 214 & 2205 & 2419 & 0.67\% & 8.85\% & 91.15\% \\ 
        editor & 317236 & 54 & 789 & 843 & 0.27\% & 6.41\% & 93.59\% \\ 
        farmer & 673603 & 278 & 1023 & 1301 & 0.19\% & 21.37\% & 78.63\% \\ 
        guard & 273362 & 2425 & 1366 & 3791 & 1.39\% & 63.97\% & 36.03\% \\ 
        hairdresser & 3399 & 4 & 15 & 19 & 0.56\% & 21.05\% & 78.95\% \\ 
        housekeeper & 3068 & 6 & 16 & 22 & 0.72\% & 27.27\% & 72.73\% \\ 
        janitor & 2701 & 5 & 1 & 6 & 0.22\% & 83.33\% & 16.67\% \\ 
        laborer & 34311 & 47 & 82 & 129 & 0.38\% & 36.43\% & 63.57\% \\ 
        lawyer & 302553 & 676 & 3030 & 3706 & 1.22\% & 18.24\% & 81.76\% \\ 
        librarian & 29699 & 12 & 134 & 146 & 0.49\% & 8.22\% & 91.78\% \\ 
        manager & 488888 & 72 & 674 & 746 & 0.15\% & 9.65\% & 90.35\% \\ 
        % mechanician & 141 & 0 & 0 & 0 & 0.00\% & NA & NA \\ 
        mover & 4151 & 5 & 0 & 5 & 0.12\% & 100.00\% & 0.00\% \\ 
        nurse & 70976 & 196 & 303 & 499 & 0.70\% & 39.28\% & 60.72\% \\ 
        physician & 174951 & 125 & 2168 & 2293 & 1.31\% & 5.45\% & 94.55\% \\ 
        receptionist & 1790 & 1 & 4 & 5 & 0.28\% & 20.00\% & 80.00\% \\ 
        salesperson & 653 & 0 & 4 & 4 & 0.61\% & 0.00\% & 100.00\% \\ 
        secretary & 358032 & 60 & 539 & 599 & 0.17\% & 10.02\% & 89.98\% \\ 
        sheriff & 63367 & 65 & 102 & 167 & 0.26\% & 38.92\% & 61.08\% \\ 
        supervisor & 34597 & 21 & 79 & 100 & 0.29\% & 21.00\% & 79.00\% \\ 
        teacher & 543672 & 306 & 1109 & 1415 & 0.26\% & 21.63\% & 78.37\% \\ 
        writer & 1247014 & 323 & 11335 & 11658 & 0.93\% & 2.77\% & 97.23\% \\ 
        \bottomrule
    \end{tabular}%
    \caption{Statistics of occupation mentions and gender-specific mentions in the \llama Wikipedia dataset.}
\end{table}
\clearpage

\section{Correlation between femaleness of an occupation and mentioning in Wikipedia is weak}\label{sec:femaleessinwiki}
    
\begin{figure}[ht]
  \centering
  \begin{subfigure}[b]{0.3\textwidth}
    \includegraphics[width=\textwidth]{images/h.png}
    \label{fig:plot1}
    \caption{}
  \end{subfigure}
  \hfill
  \begin{subfigure}[b]{0.3\textwidth}
    \includegraphics[width=\textwidth]{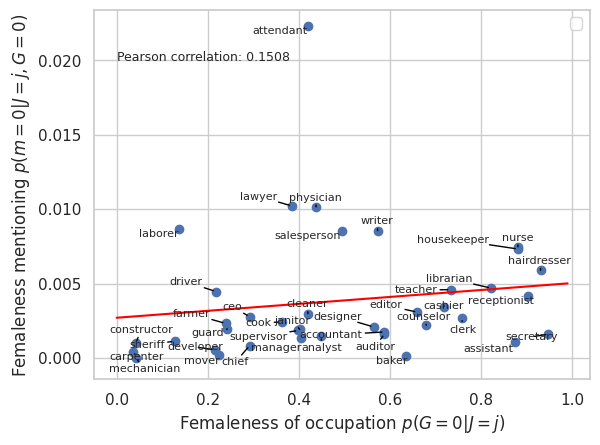}
    \label{fig:plot2}
    \caption{}
  \end{subfigure}
  \hfill
  \begin{subfigure}[b]{0.3\textwidth}
    \includegraphics[width=\textwidth]{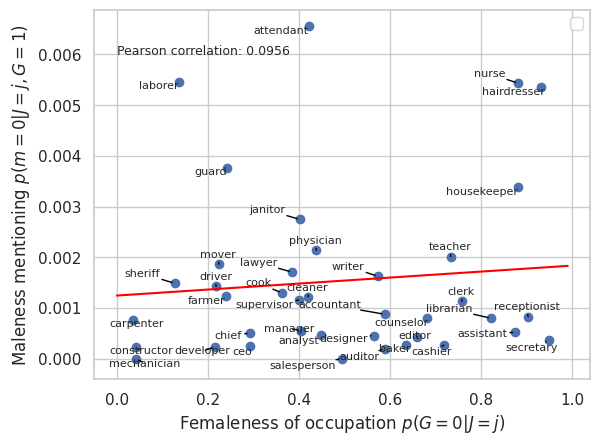}
    \label{fig:plot3}
    \caption{}
  \end{subfigure}
  \caption{Correlation between femaleness of an occupation and mentioning in Wikipedia.}
  \label{fig:salientgenderwiki}
\end{figure}
We found only weak correlations ($r\leq0.20$) between the femaleness of an occupation (according to US labor statistics) and gender, femaleness, maleness mentioning in \textbf{Wikipedia}. The strongest correlation ($r=0.20$) was for overall gender mentions.
\clearpage
\section{Female-coded and women-dominated}\label{sec:femcodewomendom}
\begin{figure}[h]
    \centering
    \includegraphics[scale=0.8]{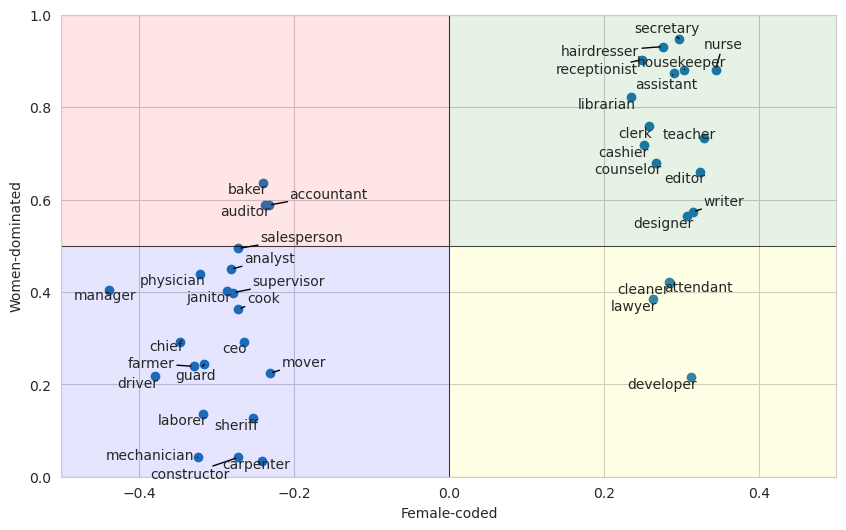}
    \caption{Plot of relationship between female-coded and women-dominated occupations.}
    \label{fig:enter-label}
\end{figure}
The y-axis, representing woman-dominatedness, is defined by the percentage of female labor representation in each occupation according to the US Bureau of Statistics. The x-axis, representing female-codedness, is the perceived femaleness of an occupation, calculated by average cosine similarity of word embedding between the occupation and two groups of gendered adjectives \footnote{[``male'', ``masculine'', ``man''] for the male group and [``female'', ``feminine'', ``woman''] for the female group.} as a proxy. Defined as \texttt{similarity\_female if similarity\_female > similarity\_male else -similarity\_male}
Occupations with a y-axis value exceeding 0.5 are considered women-dominated, while those with a positive x-axis value are deemed female-coded.

\clearpage
\section{Comparison between Pushshift.io Reddit, Wikipedia and Labor Statistics}
\label{sec:comparison}
\begin{table}[!ht]
    \centering
    \begin{tabular}{lccc}
    \toprule
        Occupation & Statistics & Wikipedia & Pushshift.io Reddit \\ 
        \midrule
        accountant & 58.80\% & 66.67\% & 42.53\% \\ 
        analyst & 44.88\% & 76.32\% & 50.77\% \\ 
        assistant & 87.48\% & 67.70\% & 59.54\% \\ 
        % attendant & 42.11\% & 77.26\% & 42.86\% \\ 
        auditor & 58.80\% & 89.29\% & 49.51\% \\ 
        baker & 63.60\% & 30.00\% & 36.40\% \\ 
        carpenter & 3.50\% & 34.88\% & 40.96\% \\ 
        cashier & 71.80\% & 92.86\% & 61.79\% \\ 
        ceo & 29.20\% & 91.46\% & 69.13\% \\ 
        chief & 29.20\% & 62.50\% & 41.14\% \\ 
        cleaner & 42.00\% & 70.97\% & 44.25\% \\ 
        clerk & 75.81\% & 70.42\% & 63.77\% \\ 
        constructor & 4.20\% & 81.82\% & 59.71\% \\ 
        cook & 36.24\% & 65.01\% & 42.26\% \\ 
        counselor & 67.97\% & 73.33\% & 53.91\% \\ 
        designer & 56.37\% & 82.03\% & 50.64\% \\ 
        developer & 21.51\% & 70.31\% & 67.38\% \\ 
        driver & 21.80\% & 75.79\% & 67.51\% \\ 
        editor & 66.00\% & 87.54\% & 51.64\% \\ 
        farmer & 23.90\% & 65.31\% & 35.51\% \\ 
        guard & 24.30\% & 34.12\% & 39.55\% \\ 
        hairdresser & 93.10\% & 52.38\% & 47.52\% \\ 
        housekeeper & 88.10\% & 68.42\% & 58.96\% \\ 
        janitor & 40.20\% & 41.18\% & 31.65\% \\ 
        laborer & 13.68\% & 61.34\% & 39.38\% \\ 
        lawyer & 38.50\% & 85.67\% & 59.61\% \\ 
        librarian & 82.20\% & 85.48\% & 49.43\% \\ 
        manager & 40.50\% & 71.05\% & 48.47\% \\ 
        % mechanician & 4.20\% & NA & NA \\ 
        mover & 22.40\% & 9.09\% & 24.41\% \\ 
        nurse & 88.09\% & 57.90\% & 32.73\% \\ 
        physician & 43.80\% & 82.60\% & 59.42\% \\ 
        receptionist & 90.30\% & 83.33\% & 55.28\% \\ 
        salesperson & 49.40\% & 100.00\% & 68.38\% \\ 
        secretary & 94.80\% & 81.66\% & 62.31\% \\ 
        sheriff & 12.70\% & 43.06\% & 35.01\% \\ 
        supervisor & 39.86\% & 61.62\% & 59.38\% \\ 
        teacher & 73.30\% & 69.50\% & 46.50\% \\ 
        writer & 57.30\% & 84.02\% & 59.83\% \\ 
        average & 46.90\% & 73.28\% & 50.96\% \\ 
        \bottomrule
    \end{tabular}
    \caption{Comparison of female representation in occupations vs. their proportional mentions in Wikipedia and Pushshift.io Reddit.}
\end{table}
\newpage
\section{Mutual information for Pushshift.io Reddit and Wikipedia}\label{sec:miredditwiki}
\begin{table}[!ht]
    \centering
    \begin{tabular}{lcc}
    \hline
        Occupations & MI Pushshift.io Reddit & MI Wikipedia \\ 
        \toprule
        accountant & 1.3E-03 & 2.1E-03 \\ 
        analyst & 3.3E-03 & 3.8E-03 \\ 
        assistant & 1.0E-02 & 1.2E-02 \\ 
        attendant & 2.0E-05 & 5.4E-06 \\ 
        auditor & 5.1E-04 & 2.7E-03 \\ 
        baker & 3.5E-03 & 1.9E-03 \\ 
        carpenter & 9.5E-03 & 3.6E-02 \\ 
        cashier & 2.2E-05 & 2.5E-04 \\ 
        ceo & 4.3E-04 & 3.2E-04 \\ 
        chief & 1.9E-05 & 7.4E-04 \\ 
        cleaner & 3.6E-06 & 1.5E-04 \\ 
        clerk & 3.8E-03 & 1.3E-02 \\ 
        constructor & 5.4E-06 & 3.1E-06 \\ 
        cook & 3.3E-05 & 1.3E-04 \\ 
        counselor & 4.9E-04 & 1.4E-04 \\ 
        designer & 1.3E-02 & 1.8E-02 \\ 
        developer & 4.4E-03 & 3.1E-03 \\ 
        driver & 1.5E-03 & 1.5E-04 \\ 
        editor & 1.3E-02 & 8.9E-02 \\ 
        farmer & 5.8E-03 & 1.8E-02 \\ 
        guard & 8.9E-04 & 7.2E-04 \\ 
        hairdresser & 1.2E-01 & 7.7E-02 \\ 
        housekeeper & 9.9E-02 & 1.5E-02 \\ 
        janitor & 7.5E-05 & 6.0E-03 \\ 
        laborer & 8.6E-06 & 9.6E-05 \\ 
        lawyer & 2.3E-04 & 5.1E-05 \\ 
        librarian & 8.6E-03 & 1.1E-02 \\ 
        manager & 1.1E-03 & 2.0E-04 \\ 
        % mechanician & 2.8E-03 & 9.2E-04 \\ 
        mover & 3.2E-04 & 1.9E-03 \\ 
        nurse & 5.9E-04 & 6.1E-04 \\ 
        physician & 2.8E-03 & 5.6E-06 \\ 
        receptionist & 5.2E-05 & 4.1E-03 \\ 
        salesperson & 8.9E-05 & 1.8E-04 \\ 
        secretary & 2.8E-04 & 2.9E-04 \\ 
        sheriff & 1.6E-02 & 8.6E-04 \\ 
        supervisor & 6.5E-03 & 4.6E-04 \\ 
        teacher & 4.8E-03 & 2.1E-02 \\ 
        writer & 5.9E-02 & 2.5E-02 \\ 
        \bottomrule
    \end{tabular}
    \caption{Computation of Mutual Information for each occupation across Pushshift.io Reddit and Wikipedia datasets.}
\end{table}
\clearpage
\section{Annotation of samples from Pushshift.io Reddit}
\begin{table*}[!ht]
\centering
\begin{tabular}{lcccc}
\hline
& Female Carpenter & Male Carpenter & Female Nurse & Male Nurse \\
\toprule
Sentiment Score & 0.4 & 0.6 & 0.3 & 0.4        \\
IsOffensive \% & 0.5 & 0.4& 0.7& 0.6        \\
IsTalkingAboutGender \%& 0.8& 0.3& 0.7& 0.6       \\
\bottomrule
\end{tabular}
    \caption{Summary of statistics from 80 Pushshift.io Reddit comments mentioning these occupations (20 each). The metrics shown are averages. Sentiment scores range from 0 to 2 (0 = negative, 1 = neutral, 2 = positive), obtained using Stanford Stanza sentiment analysis model. The categories `IsOffensive' and `IsTalkingAboutGender' are annotated by the authors using a binary system (0 = No, 1 = Yes), with the percentages indicating the average frequency of `Yes' responses.}\label{tab:80comments}
\end{table*}

\begin{figure*}[ht]
  \centering
  \begin{subfigure}[b]{0.5\linewidth}
    \centering
    \includegraphics[width=\linewidth]{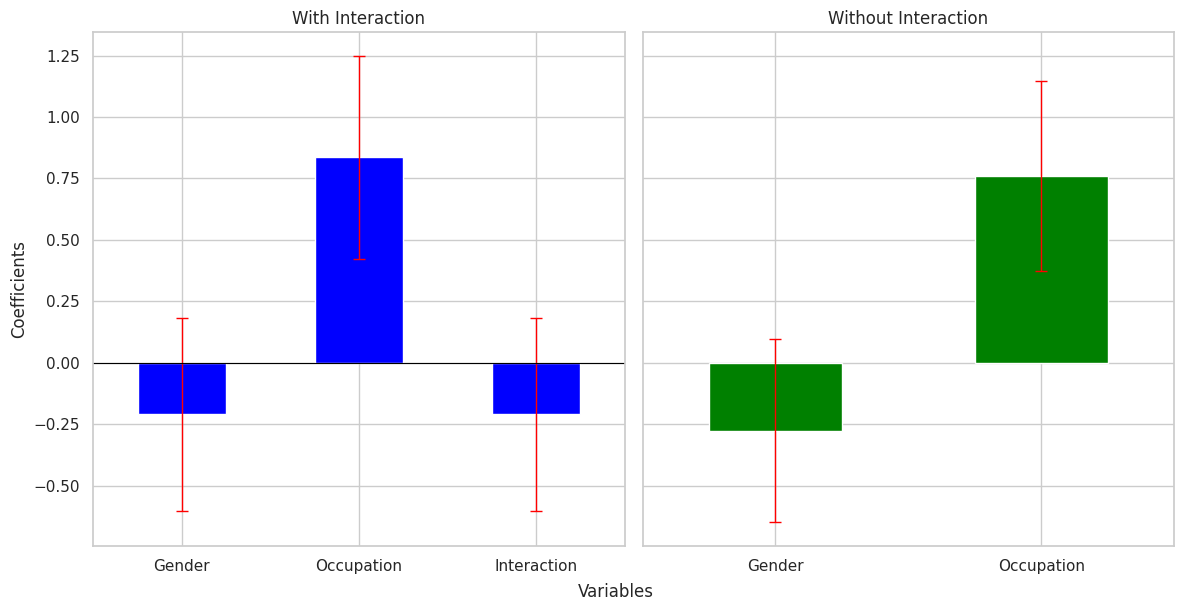}
    \caption{Regression coefficients predicting IsOffensive}
    \label{fig:reddit_annotation_offensive}
  \end{subfigure}
  \hfill
  \begin{subfigure}[b]{0.5\linewidth}
    \centering
    \includegraphics[width=\linewidth]{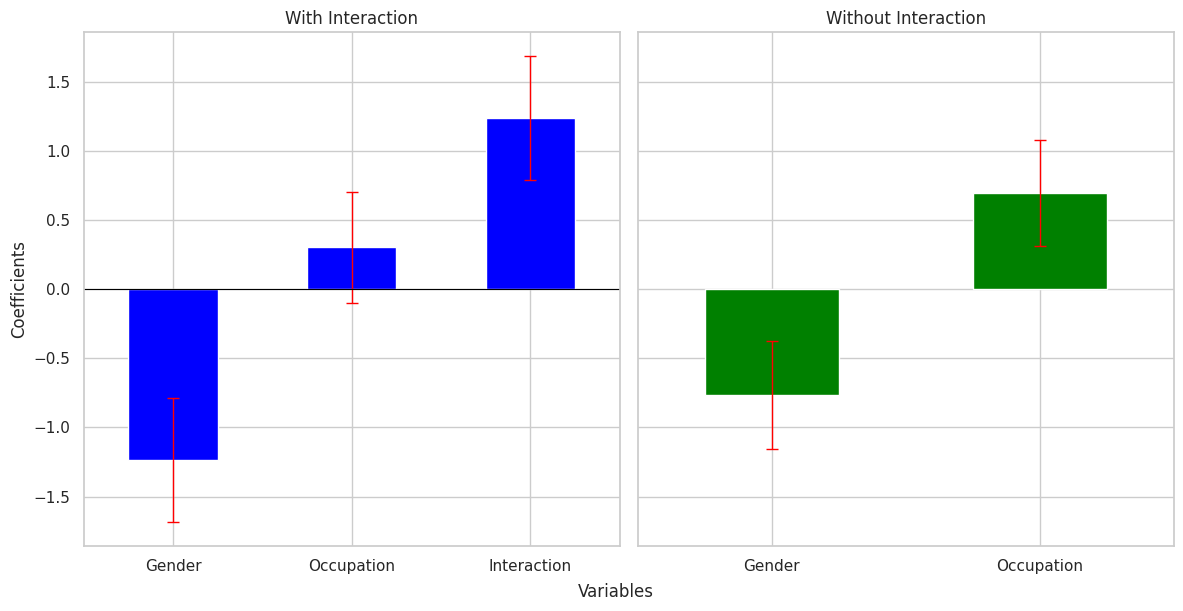}
    \caption{Regression coefficients predicting IsTalkingAboutGender}
    \label{fig:reddit_annotation_aboutgender}
  \end{subfigure}
  \caption{In our study, we used an XNOR operation to create an interaction variable between gender (Female = 0, Male = 1) and occupation (Carpenter = 0, Nurse = 1), identifying non-typical roles (Male Nurses and Female Carpenters) as 1. The interaction has minimal impact in example (a) but is crucial in example (b). Figure \ref{fig:reddit_annotation_offensive} reveals more offensive language in discussions about female-coded roles like ``nurse'', while Figure \ref{fig:reddit_annotation_aboutgender} shows that non-typical gender roles are often highlighted in gender-related conversations.}
  \label{fig:salientgender}
\end{figure*}
% \begin{table*}[!ht]
% \centering
% \begin{tabular}{lcc}
% \hline
% Test                        & Chi-squared Statistic & p-value \\
% \toprule
% Gender vs IsOffensive       & 0.05                  & 0.82    \\
% Occupation vs IsOffensive   & 2.47                  & 0.12    \\
% Gender vs IsAboutGender     & 6.11                  & 0.01    \\
% Occupation vs IsAboutGender & 0.05                  & 0.82    \\
% \bottomrule
% \end{tabular}
%     \caption{Summary of Chi-squared statistic results.}
% \end{table*}

\end{document}